\pdfoutput=1

\documentclass[11pt]{article}

\usepackage[preprint]{acl}

\usepackage{times}
\usepackage{latexsym}

\usepackage[T1]{fontenc}

\usepackage[utf8]{inputenc}

\usepackage{microtype}

\usepackage{inconsolata}

\usepackage{graphicx}

\usepackage{hyperref}       
\usepackage{url}            
\usepackage{booktabs}       
\usepackage{amsfonts}       
\usepackage{nicefrac}       
\usepackage{xcolor}         
\usepackage{amsmath}
\usepackage{enumitem}
\usepackage{makecell} 
\usepackage{algorithm}
\usepackage{algorithmic}

\usepackage{multirow}
\usepackage{soul}
\usepackage{subcaption}
\usepackage{bm}
\usepackage{wasysym}
\usepackage{pifont}
\newcommand{\cmark}{\ding{51}}%
\newcommand{\xmark}{\ding{55}}%

\title{LLM-ML Teaming: Integrated Symbolic Decoding and Gradient \\ Search for Valid and Stable Generative Feature Transformation}

\author{
\textbf{Xinyuan Wang\textsuperscript{1}},
\textbf{Haoyue Bai\textsuperscript{1}},
\textbf{Nanxu Gong\textsuperscript{1}},
\textbf{Wangyang Ying\textsuperscript{1}},
\\
\textbf{Sixun Dong\textsuperscript{1}},
\textbf{Xiquan Cui\textsuperscript{2}},
\textbf{Yanjie Fu\textsuperscript{1}}
\\
\\
\textsuperscript{1}Arizona State University,
\textsuperscript{2}The Home Depot
\\
\{xwang735, haoyuebai, ngong6, wying4, sixundong, yanjie.fu\}@asu.edu \\ xiquan\_cui@homedepot.com
}

\begin{document}
\maketitle

\begin{abstract}

Feature transformation enhances data representation by deriving new features from the original data. 
Generative AI offers potential for this task, but faces challenges in stable generation (consistent outputs) and valid generation (error-free sequences). 
Existing methods—traditional ML’s low validity and LLMs’ instability—fail to resolve both. 
We find that LLMs ensure valid syntax, while ML’s gradient-steered search stabilizes performance. 
To bridge this gap, we propose a teaming framework combining LLMs’ symbolic generation with ML’s gradient optimization. 
This framework includes four steps: (1) golden examples generation, aiming to prepare high-quality samples with the ground knowledge of the teacher LLM; (2) feature transformation sequence embedding and search, intending to uncover potentially superior embeddings within the latent space; (3) student LLM feature transformation, aiming to distill knowledge from the teacher LLM; (4) LLM-ML decoder teaming, dedicating to combine ML and the student LLM probabilities for valid and stable generation.
The experiments on various datasets show that the teaming policy can achieve 5\% improvement in downstream performance while reducing nearly half of the error cases. The results also demonstrate the efficiency and robustness of the teaming policy. Additionally, we also have exciting findings on LLMs' capacity to understand the original data.
The codes are available at \href{https://anonymous.4open.science/r/LLM-ML-Teaming-DCAI-B391}{this link}.

\end{abstract}

\section{Introduction}
Feature transformation is to derive a new feature set from an original feature set to reprogram data representation, for instance, transforming $[a, b]$ into $[a/b, a-b, (a+b)/a]$. 
Feature transformation can reconstruct distance measures, reshape discriminative patterns, and enhance data AI readiness (e.g., structural, predictive, interaction, and expression levels). 
Generative AI (e.g., LLM) has the potential to deliver far better features~\cite{zhao2023survey} than manual reconstruction or machine-assisted approaches (e.g., genetic algorithms, simulated annealing, reinforcement learning).
Generative Feature Transformation (GFT) formulates the task as a sequence generation problem, where each transformed feature (e.g., $a/b$) is treated as a token, and a new feature set (e.g., $[a/b, a-b, (a+b)/a]$) becomes a token sequence~\cite{wang2025towards,ying2025survey}. Solving GFT efficiently avoids exhaustive search over exponentially large spaces and accelerates automated feature engineering.

\begin{figure}[t]
    \centering
    \begin{subfigure}[b]{0.20\textwidth}
        \includegraphics[width=\textwidth]{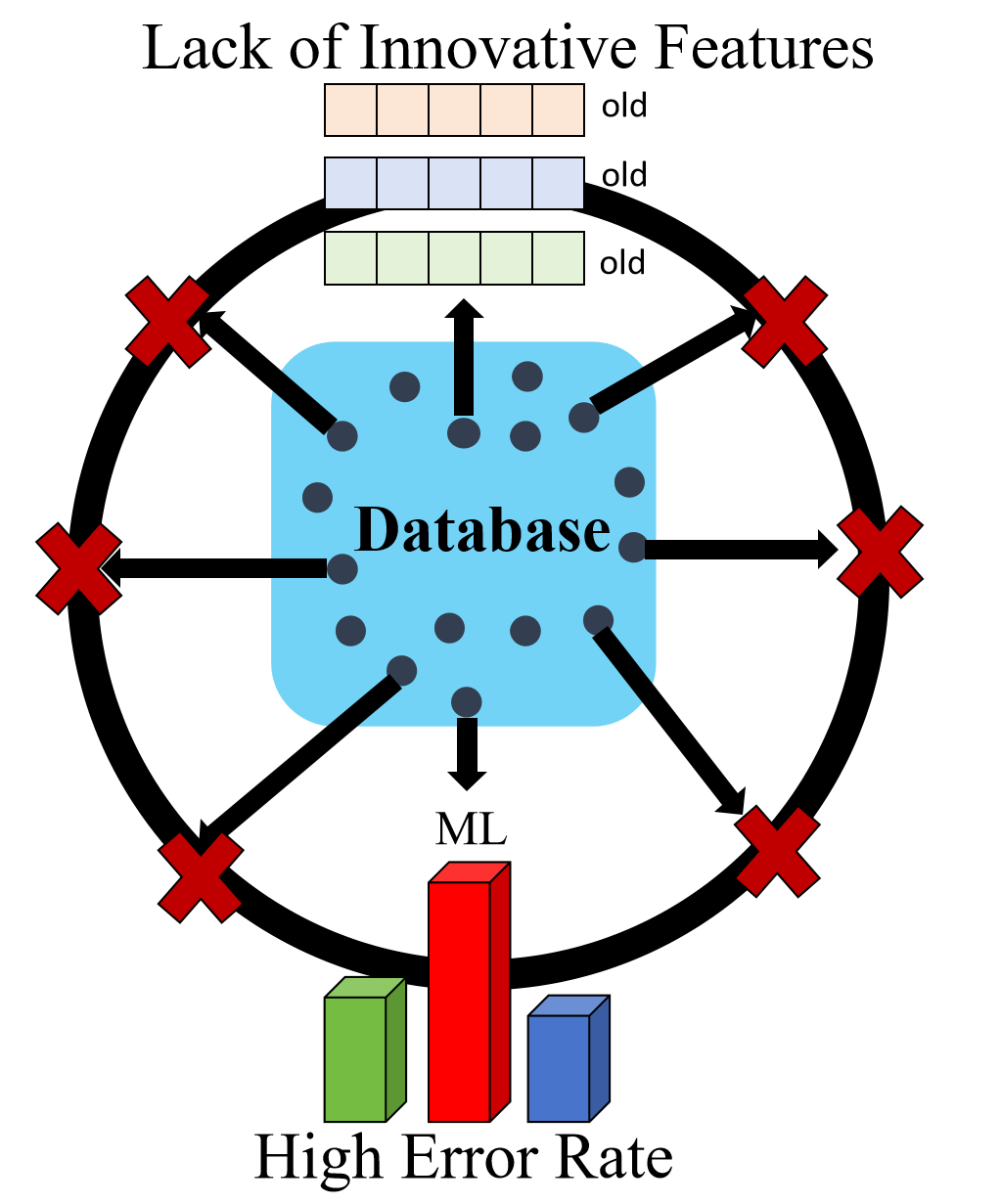}
        \caption{ML Drawbacks}
        \label{fig:intro_drawback_a}
    \end{subfigure}
    \hfill
    \begin{subfigure}[b]{0.24\textwidth}
        \includegraphics[width=\textwidth]{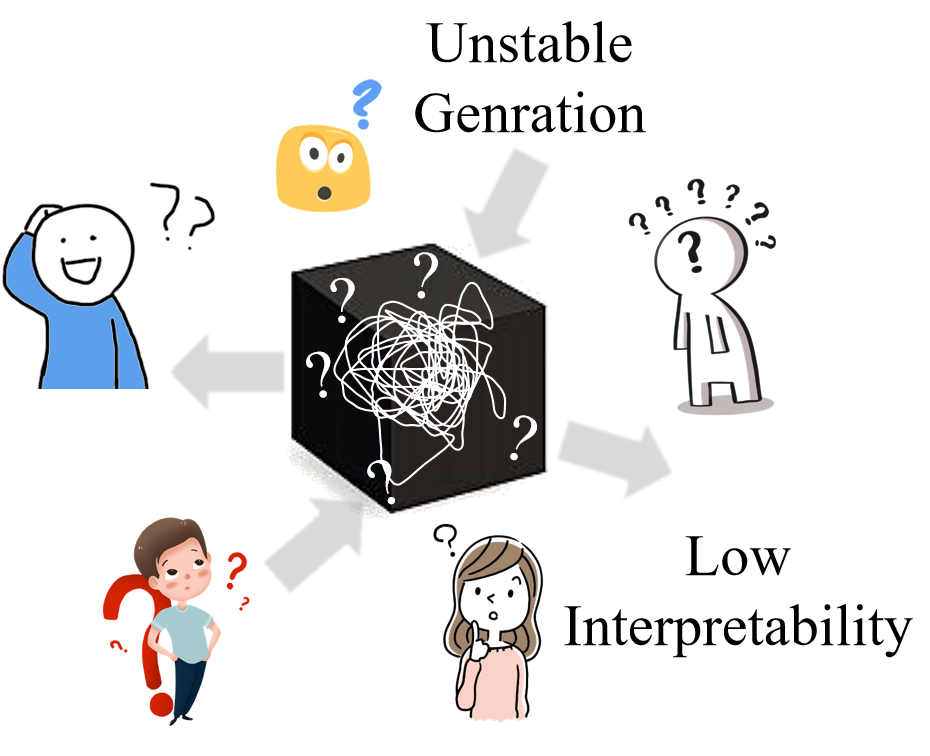}
        \caption{LLMs Drawbacks}
        \label{fig:intro_drawback_b}
    \end{subfigure}
    \vspace{-0.3cm}
    \caption{Traditional ML’s Low Validity and LLMs’ Instability.}
    \label{fig:intro_drawback_both}
\end{figure}

There are two major challenges (\textbf{Figure~\ref{fig:intro_drawback_both}}) in solving GFT: (1) stable generation, and (2) valid generation.
First, some generative methods exhibit unstable variability by causing significant shifts in generated features; that is, the same input feature set results in different feature transformations across different runs with different performances. 
Stable generation seeks to answer: how can we ensure the consistency and stability of generated features across different inputs and runs?
Second, we observed that some generative methods can generate undefined values (e.g., division by zero), violate mathematical constraints, and introduce redundancy that does not contribute to feature transformations. 
Valid generation of feature transformations is intended to answer: how can we generate legal and sound feature transformation sequences that contribute to downstream performance?

Relevant work can only partially solve the two challenges of GFT.
First, GFT is related to automated discrete search-based approaches, such as genetic algorithms, simulated annealing, and reinforcement learning, which search for optimal feature transformations.
However, these methods suffer from an exponential search space and rely on hand-crafted reward functions.
Second, GFT is connected to deep sequential learning  (e.g., encoder-decoder architectures), which aims to learn the data embedding and decode the embedding into feature transformation sequences.
However, such methods often generate illegal tokens due to the lack of robust tokenization, post-validation mechanisms, and weak syntax enforcement.
Third, GFT is related to LLMs, where we fine-tune models to generate feature transformation sequences.
However, LLMs suffer from instability and the preference for using simple operators (e.g., addition)~\cite{kuken2024large} to generate different feature transformations with different performances in different runs, due to stochastic sampling and probabilistic token selection.
Existing studies demonstrate the inability to jointly address both stable and valid generation in GFT.
As a result, a new method is needed to achieve stable and valid feature transformations.

\textbf{Our Perspective: teaming ML gradient search for stability and LLM symbolic generation for validity.}
After a massive analysis, we have two observations:
(1) While LLMs can generate different feature transformation sequences with different performances across different runs, LLMs are capable of generating valid, legal token expressions of feature transformation;
(2) another solution is encoding-search-decoding which computes the embedding space of data to transform, then leverages gradient search to identify the best embedding space, and decodes the best embedding space into optimal feature transformation sequence. While such the method generates illegal tokens, its gradient search can ensure that the identified embedding of feature transformation is better than initialization embeddings, thus, demonstrating stable performance improvements.  
We derive two key insights from the two observations: (1) LLM symbolic generation for valid generation; and (2) ML gradient search for stable generation. 
Our perspective is to team LLM symbolic generation with  ML gradient search together to achieve a valid and stable generation of feature transformations. 
We highlight that leveraging teacher LLM-generated data to train both the ML model and the student LLM, along with collaboratively decoding between the student LLM and ML decoder, is an effective way to integrate both validity and stability.
Our work's key innovation focuses on LLM–ML teaming rather than solely relying on LLMs. Our method incorporates an ML gradient-based search with LLM symbolic generation to address LLM instability and ML validity limitations. 

\textbf{Summary of Proposed Solution: }
Inspired by these findings, we develop a four-step LLM-ML teaming framework to integrate valid symbolic generation and stable gradient-steered search. 
\textit{Step 1 data} is to leverage generic LLM (i.e., ChatGPT-4 API and prompting) to generate high-quality and diverse transformed feature sets, along with corresponding performance on a downstream task (e.g., random forest classification) as golden training examples. 
\textit{Step 2 stability} is to exploit the golden training examples of Step 1 to train an embedding, gradient-steered search, decoding based ML pipeline for GFT. The gradient-steered mountain-climbing search provides stable improvements in identifying better feature transformations in an embedding space. 
\textit{Step 3 validity} is to utilize the gold training examples of Step 1 to fine-tune a foundation LLM model with subword mechanism, contextual self-attention, and structured data pre-training through two tasks:  sequence reconstruction and corresponding feature performance prediction. This is to build the LLM side with logits for teaming.
\textit{Step 4: collaboration} is to integrate stable search in ML and valid generation in LLM by calibrating LLM's next token probability using the next token probability of the gradient search based decoder.
Extensive experiments show that the teaming of ML's gradient search and LLM's generation can improve the validity and stability of GFT. In addition, it achieves 5\% improvement on such generalized and challenging feature engineering tasks.  

\textbf{Our Contributions:}
(1) Formulation: We tackle an interesting problem: stability and validity in generative feature transformation, which is an automated data engineering task. 
(2) Insights: we find that gradient-steered search can strengthen generation stability on performance improvement in GFT; LLMs' symbolic generation can improve valid and legal generation. 
(3) Techniques: we propose an LLM-ML teaming strategy to integrate valid symbolic generation and stable gradient-steered search. 
The integration is achieved through teacher-guided training and collaborative decoding.

\section{Preliminaries and Problem Statement}

\subsection{Important Concepts}
\label{sec:concept}

\noindent\textbf{Operation Set.} We define a set of mathematical operations, including unary (e.g., \texttt{log}, \texttt{exp}) and binary (e.g., \texttt{add}, \texttt{divide}) operators. The operators are applied to existing features to construct new ones.

\begin{figure}[htbp] 
    \centering 
    \includegraphics[width=0.45\textwidth]{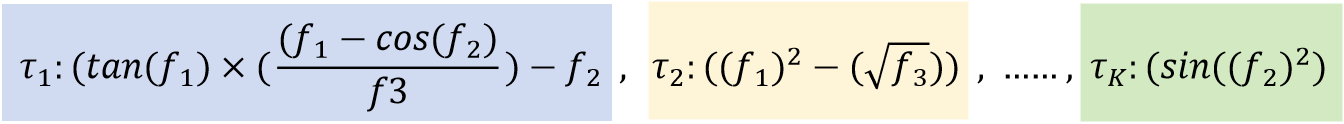} 
    \caption{A Feature Transformation Sequence Example.} 
    \label{fig:example_seq}
\end{figure}

\noindent\textbf{Feature Transformation Sequence.} A feature transformation sequence is a collection of symbolic expressions that define how raw features are combined. These expressions are represented as token sequences composed of feature IDs and operators. \textbf{Figure~\ref{fig:example_seq}} shows an example.

\noindent\textbf{Postfix Representation.} To reduce ambiguity and simplify decoding, we adopt the postfix notation instead of infix. Postfix sequences eliminate the need for brackets and enable left-to-right parsing. \textbf{Figure~\ref{fig:example_postfix_all}} illustrates the difference. See \textbf{Appendix~\ref{app:concept}} for examples and details.

\subsection{Problem Statement}
We aim to develop a generative AI system that generates a feature transformation sequence given a tasking dataset, by integrating LLM symbolic generation for valid generation and ML gradient search for stable generation. Formally, given a dataset $D = \{X,y\}$ and an operation set $\mathcal{O}$, the goal is to find the optimal feature transformation sequence $\Gamma^{*}$ that maximizes the downstream ML model $\mathcal{M}$'s performance (i.e.,  balance among accuracy, validity, and stability) on the transformed feature set:
\begin{equation}
\Gamma^{*} = \underset{\Gamma}{\operatorname{argmax}} \mathcal{A}(\mathcal{M}(\text{Transform}(X, \Gamma)), y)
\end{equation}
where $\text{Transform}(X, \Gamma)$ transforms the original feature set $X$ using $\Gamma$, and $\mathcal{A}$ is the downstream performance metric for $\mathcal{M}$. 

\begin{figure*}[htbp]
    \centering
    \includegraphics[width=0.99\textwidth]{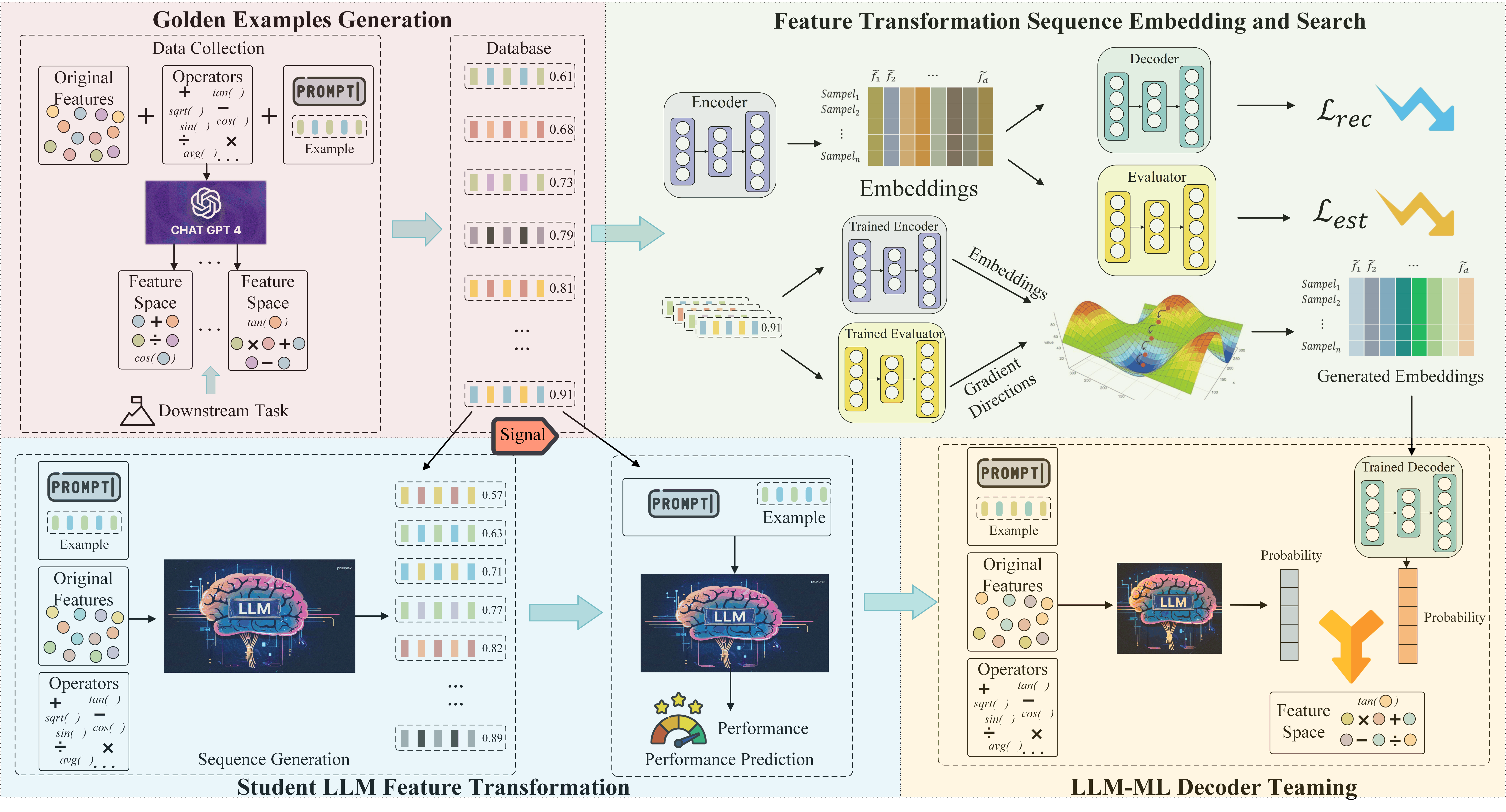}
    \caption{Overview of the LLM-ML Teaming Framework. The framework consists of four key components: (1) Golden Examples Generation, where LLMs generate high-quality feature transformation examples; (2) Feature Transformation Sequence Embedding and Search, optimizing transformation sequences in the latent space; (3) Student LLM Feature Transformation, distilling knowledge into a compact LLM; and (4) LLM-ML Decoder Teaming, refining sequence generation via teaming decoding.}

    \label{fig:method_overview}
\end{figure*}

\section{Proposed Method}

\subsection{Framework Overview}

\textbf{Figure~\ref{fig:method_overview}} shows the framework includes four components:
(1) \textbf{Golden Examples Generation.} We use an advanced LLM (e.g., ChatGPT) to generate feature transformation sequences and design downstream  tasks (e.g., regression or classification) to evaluate each sequence. The sequences and their evaluation scores form golden examples that serve as training data.
(2) \textbf{Transformation Sequence Embedding and Search.} 
To learn golden examples and facilitate the exploration of the optimal transformation path, we deploy an  encoder-evaluator-decoder ML framework. The encoder embeds feature transformation sequences into fixed-length vectors to construct an embedding space. 
The evaluator assesses the utility of these embedding vectors of feature transformations and provides gradient guidance to search for optimal embeddings in the latent embedding space. The decoder reconstructs embeddings into transformation sequences.
(3) \textbf{LLM Supervised Fine-tuning.} Since ChatGPT is a black-box model that only outputs discrete tokens, we cannot directly access its probability distribution. Additionally, ChatGPT is too large to fine-tune efficiently. Therefore, we adopt a lightweight Llama model as a student LLM to learn from the golden examples. This allows the student LLM to become more efficient and compact while acquiring knowledge of feature transformations and generating transformation sequences with probabilistic outputs.
(4) \textbf{LLM-ML Decoder Teaming.} We leverage LLM-generated probability to guide the ML model's decoding process. 
The ML decoder, informed by the finetuned LLM’s prior knowledge, improves the stability and validity of the results.

\subsection{Golden Examples from Teacher LLM}
We leverage powerful APIs, such as GPT-4o, to generate golden examples as training data. 
Given a dataset $D = \{X, y\}$ with the features $X = [f_{1}, \cdots, f_{N}]$, we construct prompts that cross original features with operators from the operation set $\mathcal{O}$. Besides, we add certain rules and one-shot example about how to transform a feature set in a prompt, to guide the LLM.  A sample prompt is described in \textbf{Appendix~\ref{app:prompt}}.
Under such prompts, the LLM generates feature transformation sequences, thereafter evaluated on downstream tasks to obtain performance. 
The resulting pairs form a high-quality database, denoted as $(\Gamma_i, s_i)_{i=1}^M$, where $\Gamma$ and $s$ represent the feature transformation sequence and the downstream performance, respectively, and $M$ is the number of golden examples. This high-quality database serves two purposes: (1) distilling the knowledge of the teacher LLM as a reference for fine-tuning the student LLM, enabling it to generate more efficient and controlled feature transformations, and (2) providing diverse, high-performing samples to guide search-based ML methods, ensuring that the optimization process explores a well-informed and promising search space while avoiding suboptimal or redundant transformations.

Golden examples provide high-quality training data and optimization signals to guide the directions of optimal feature transformation search and generation. 
Compared to random methods, golden examples help narrow the search space, making the optimization process more efficient. The evaluator assesses golden examples, steering the search toward high-quality feature transformation patterns while avoiding ineffective exploration. Additionally, golden examples establish an experience-driven search boundary, allowing search methods to focus on meaningful paths, thereby improving the accuracy and stability of generations.
Golden examples provide a strong foundation for guiding both the search process and the student LLM, but their quality and diversity depend on the generation method. Traditional RL-based algorithms often struggle to generate innovative feature crosses, as they tend to follow fixed reward patterns, leading to repetitive and predictable outputs. In contrast, LLMs, with their vast general knowledge, can generate a wide variety of feature crosses, including unconventional and innovative patterns that traditional methods may overlook. By leveraging LLMs to generate golden examples, we ensure a richer and more diverse set of high-quality transformation sequences, which in turn enhances the effectiveness of the search process and encodes more knowledge into the student LLM.

\subsection{Feature Transformation Sequence Embedding and Search}

This module is designed to explore and optimize feature transformation sequence embeddings within a latent space. It employs a gradient-steered search approach to iteratively refine transformation sequences, ensuring that the generated features align closely with the objectives of downstream tasks. The module adopts an encoder-evaluator-decoder structure: the encoder maps transformation sequences into a high-dimensional latent space, the evaluator predicts downstream performance based on embeddings, and the decoder reconstructs embeddings back into transformation sequences.

\subsubsection{Training of Encoder-Evaluator-Decoder}

To build a robust encoder-evaluator-decoder structure, a multi-step training strategy is employed, leveraging the golden examples as reference.
We use a bidirectional GRU encoder and a two-layer MLP evaluator. The decoder is the LSTM followed by a token classifier.

Given a transformation sequence $\Gamma = [\tau_1, \tau_2, \dots, \tau_K]$, the encoder maps it to a latent embedding $\mathbf{z}_i = \text{Encoder}(\Gamma_i)$.
The latent embedding $\mathbf{z}_i$ is then fed into both the decoder and the evaluator, which are trained jointly. 
The decoder minimizes the reconstruction loss $\mathcal{L}_{\text{rec}} = \| \Gamma_i - \text{Decoder}(\text{Encoder}(\Gamma_i)) \|_2^2$ to ensure the embeddings retain sufficient information to reconstruct the original sequence.
The evaluator predicts the downstream performance of each sequence based on the golden examples with the prediction loss $\mathcal{L}_{\text{est}} = \frac{1}{N} \sum_{i=1}^N \left( \hat{s}_i - s_i \right)^2$, 
where $\hat{s}_i$ is the predicted performance, and $s_i$ is the corresponding ground-truth value.
The joint training objective combines these two losses: $\mathcal{L}_{\text{joint}} = \alpha \mathcal{L}_{\text{rec}} + ( 1 - \alpha )\mathcal{L}_{\text{est}}$.

\subsubsection{Embedding Search}

After training the encoder-evaluator-decoder structure, an embedding search is performed to identify high-performing transformation sequences.

First, the latent embedding $\mathbf{z}_i$ for a given sequence $\Gamma$ is obtained. 
Then the evaluator predicts the downstream performance $\hat{s}_i = \text{Evaluator}(\mathbf{z}_i)$.
The evaluator computes the performance score's gradient $\nabla_{\mathbf{z}_i} \hat{s}_i = \frac{\partial \hat{s}_i}{\partial \mathbf{z}_i}$ with respect to the embedding, guiding the search process.
Then the embedding is updated iteratively as $\mathbf{z}_i^{\text{new}} = \mathbf{z}_i + \eta \nabla_{\mathbf{z}_i} \hat{s}_i$ to maximize the predicted performance,
where $\eta$ is the learning step size.
The updated embedding $\mathbf{z}_i^{\text{new}}$ is decoded into a new transformation sequence $\Gamma_i^{\text{new}} = \text{Decoder}(\mathbf{z}_i^{\text{new}})$.

This search process iteratively alternates between embedding optimization and sequence decoding, progressively refining the quality of the generated sequences. By aligning the sequences with task objectives and exploring diverse regions of the feature space, this approach maximizes both performance and innovation.

While neither LLMs nor black-box ML models are inherently interpretable, ML-guided search offers clearer gradient-driven rationale for transformation selection. Unlike autoregressive LLM generation, which is sensitive to decoding temperature and sampling noise, ML-guided latent optimization offers smoother, reproducible search dynamics, enabling stable feature discovery.

\subsection{Student LLM Feature Transformation}
The student LLM is fine-tuned using golden samples to get the token probability for use in the decoder teaming process. Two key tasks are involved: (1) \textbf{Sequence Generation Task}, where the LLM generates transformation sequences based on input prompts to enable the student LLM to learn the structure and syntax of transformation sequences from the teacher LLM's examples, and (2) \textbf{Performance Prediction Task}, where the LLM predicts the downstream performance of the generated sequences to enhance the student model’s ability to generate valid and informative sequences.

Details of the training objectives and loss functions are provided in \textbf{Appendix~\ref{app:distillation}}.

\begin{table*}[htbp]
  \begin{center}
    \caption{Overall Downstream Performance Comparison.}
    \label{tab:overallresult}
        \resizebox{\textwidth}{!}{\begin{tabular}{lcccccccccccccccccccc}
            \toprule
            \textbf{Dataset} & \textbf{Source} & \textbf{Task} & \textbf{Samples} & \textbf{Features} & \textbf{Original} & \textbf{RDG} & \textbf{LDA} & \textbf{ERG} & \textbf{NFS} & \textbf{AFAT} & \textbf{PCA} & \textbf{TTG} & \textbf{GRFG} & \textbf{MOAT} & \textbf{FeatLLM} & \textbf{CAAFE} & \textbf{AutoFeat} & \textbf{OpenFE} & \textbf{ELLM-FT} & \textbf{Teaming} \\
            \midrule
            Amazon Employee & Kaggle & C & 32,769 & 9 & 93.37\% & 92.31\% & 91.64\% & 92.43\% & 93.21\% & 92.97\% & 92.29\% & 92.79\% & 93.02\% & 93.13\% & \textbf{93.62\%} & 91.41\% & 93.29\% & 93.44\% & 93.17\% & \ul{93.52\%} \\
            AP-omentum-ovary & OpenML & C & 275 & 10,936 & 78.16\% & 74.32\% & 59.46\% & 73.65\% & 75.00\% & 74.32\% & 73.65\% & 68.24\% & 76.35\% & 79.64\% & 78.89\% & 78.16\% & 77.63\% & 78.16\% & \ul{80.06\%} & \textbf{81.39\%} \\
            SpectF & UCIrvine & C & 267 & 44 & 76.06\% & 76.03\% & 66.29\% & 75.66\% & 79.40\% & 76.03\% & 70.92\% & 76.03\% & 81.65\% & \ul{86.95\%} & 80.07\% & 70.60\% & 76.06\% & 76.06\% & 86.14\% & \textbf{90.49\%} \\
            German Credit & UCIrvine & C & 1,000 & 24 & 74.20\% & 68.01\% & 63.91\% & 74.43\% & 68.67\% & 68.32\% & 67.92\% & 64.51\% & 68.29\% & 72.44\% & 76.35\% & 59.92\% & 74.86\% & 74.50\% & \ul{76.39\%} & \textbf{85.32\%} \\
            UCI Credit & UCIrvine & C & 30,000 & 23 & 79.29\% & 80.32\% & 74.37\% & 80.16\% & 80.13\% & 80.32\% & 73.27\% & 79.81\% & 80.67\% & \textbf{80.87\%} & 76.39\% & 76.80\% & 79.72\% & 80.11\% & 79.29\% & \ul{80.86\%} \\
            Spam Base & UCIrvine & C & 4,601 & 57 & 94.53\% & 90.61\% & 88.89\% & 91.70\% & 92.50\% & 91.20\% & 81.66\% & 91.91\% & 92.20\% & 92.90\% & 95.03\% & 88.51\% & \ul{94.54\%} & 94.53\% & \textbf{96.68\%} & 93.46\% \\
            Ionosphere & UCIrvine & C & 351 & 34 & 93.37\% & 91.17\% & 65.53\% & 92.02\% & 91.17\% & 92.87\% & 92.87\% & 90.31\% & 93.16\% & 95.69\% & 95.38\% & 92.84\% & 93.37\% & 93.37\% & \ul{96.01\%} & \textbf{97.10\%} \\
            Higgs Boson & UCIrvine & C & 50,000 & 28 & 69.66\% & 67.51\% & 51.32\% & 69.02\% & 69.17\% & 69.70\% & 53.45\% & 68.99\% & 69.77\% & 69.12\% & \ul{70.35\%} & 61.26\% & 67.35\% & 69.66\% & 69.66\% & \textbf{70.81\%} \\
            PimaIndian & Kaggle & C & 768 & 8 & 80.68\% & 76.04\% & 63.80\% & 76.17\% & 74.87\% & 76.56\% & 63.80\% & 74.48\% & 75.39\% & 80.73\% & 89.66\% & 79.86\% & 80.86\% & 80.86\% & \ul{89.66\%} & \textbf{91.95\%} \\
            Messidor Feature & UCIrvine & C & 1,151 & 19 & 69.09\% & 62.38\% & 47.52\% & 66.90\% & 63.77\% & 66.55\% & 67.21\% & 66.46\% & 69.24\% & 73.02\% & 72.62\% & 66.10\% & 69.08\% & 69.09\% & \ul{74.80\%} & \textbf{75.61\%} \\
            Wine Quality Red & UCIrvine & C & 999 & 11 & 60.95\% & 46.65\% & 43.31\% & 46.10\% & 46.21\% & 48.05\% & 42.21\% & 46.71\% & 47.01\% & 62.10\% & 62.65\% & 51.74\% & \ul{62.52\%} & 53.71\% & 61.11\% & \textbf{62.94\%} \\
            Wine Quality White & UCIrvine & C & 4,898 & 11 & 54.75\% & 52.41\% & 44.94\% & 51.04\% & 52.51\% & 51.67\% & 43.01\% & 53.12\% & 53.41\% & 54.52\% & \textbf{56.87\%} & 42.82\% & 54.26\% & 54.75\% & 55.03\% & \ul{55.18\%} \\
            SVMGuide3 & LibSVM & C & 1,243 & 21 & 81.85\% & 78.68\% & 65.24\% & 82.62\% & 79.16\% & 79.49\% & 67.60\% & 79.81\% & 81.17\% & 81.74\% & 82.54\% & 75.30\% & \ul{83.05\%} & 81.85\% & 82.70\% & \textbf{84.64\%} \\
            Lymphography &  UCIrvine & C & 148 & 18 & 83.19\% & 79.36\% & 70.38\% & 83.73\% & 85.25\% & 82.38\% & 70.38\% & 82.38\% & 85.51\% & 88.38\% & 85.24\% & 75.00\% & 79.26\% & 83.73\% & \ul{90.54\%} & \textbf{91.89\%} \\
            \midrule
            Airfoil & UCIrvine & R & 1,503 & 5 & 0.5749 & 0.5193 & 0.2201 & 0.5193 & 0.5193 & 0.5210 & 0.2730 & 0.5003 & 0.5587 & 0.5967 & 0.5877 & N/A & 0.5746 & 0.5746 & \ul{0.6174} & \textbf{0.6329} \\
            Housing Boston & Kaggle & R & 506 & 13 & 0.4148 & 0.4043 & 0.0201 & 0.4090 & 0.4251 & 0.4161 & 0.1048 & 0.3967 & 0.4043 & 0.4463 & 0.4442 & N/A & 0.4149 & 0.4148 & \ul{0.4564} & \textbf{0.4584} \\
            Openml 586 & OpenML & R & 1,000 & 25 & 0.6311 & 0.5681 & 0.1109 & 0.6147 & 0.5443 & 0.5435 & 0.1109 & 0.5443 & 0.5768 & 0.6251 & \ul{0.6477} & N/A & 0.6329 & 0.6311 & 0.6328 & \textbf{0.6569} \\
            Openml 589 & OpenML & R & 1,000 & 25 & 0.5388 & 0.5091 & 0.0112 & 0.5103 & 0.5053 & 0.5087 & 0.0112 & 0.5032 & 0.5047 & 0.5139 & 0.5545 & N/A & 0.5423 & 0.5388 & \ul{0.5836} & \textbf{0.5990} \\
            Openml 607 & OpenML & R & 1,000 & 50 & \ul{0.6207} & 0.5208 & 0.1071 & 0.5553 & 0.5194 & 0.5158 & 0.1071 & 0.5222 & 0.6021 & \ul{0.6051} & 0.5608 & N/A & 0.6191 & \textbf{0.6207} & 0.6089 & 0.6181 \\
            Openml 616 & OpenML & R & 500 & 50 & 0.3736 & 0.0701 & 0.0241 & 0.1937 & 0.1667 & 0.1489 & 0.0242 & 0.1567 & 0.3722 & 0.4063 & 0.3836 & N/A & 0.3924 & 0.3736 & \textbf{0.4082} & \ul{0.4073} \\
            Openml 618 & OpenML & R & 1,000 & 50 & 0.4402 & 0.3720 & 0.0521 & 0.3561 & 0.3473 & 0.2472 & 0.1016 & 0.3467 & 0.4562 & \ul{0.4734} & 0.4597 & N/A & 0.4407 & 0.4402 & 0.4734 & \textbf{0.4840} \\
            Openml 620 & OpenML & R & 1,000 & 25 & 0.6434 & 0.5111 & 0.0293 & 0.5466 & 0.5130 & 0.5267 & 0.1138 & 0.5123 & 0.5591 & 0.5722 & 0.5725 & N/A & \textbf{0.6576} & \ul{0.6434} & 0.6203 & 0.5847 \\
            Openml 637 & OpenML & R & 500 & 50 & \ul{0.3162} & 0.1364 & 0.0433 & 0.1521 & 0.1521 & 0.1758 & 0.0352 & 0.1439 & 0.2071 & 0.2125 & 0.2945 & N/A & 0.3251 & \textbf{0.3162} & 0.2946 & 0.3095 \\
            \midrule
            \textbf{Average Ranking} & - & - & - & - & 5.52 & 11.35 & 15.00 & 9.74 & 9.78 & 9.87 & 14.00 & 11.65 & 7.83 & 5.04 & 4.22 & 14.00 & 5.52 & 5.13 & 3.35 & 1.83 \\
            \bottomrule
        \end{tabular}}
    \end{center}
\end{table*}

\subsection{LLM-ML Decoder Teaming}

The decoder teaming policy enhances sequence generation by ensuring validity, coherence, and logical consistency. It integrates the \textbf{ML Decoder} and \textbf{LLM Decoder} using a probabilistic framework that leverages their complementary strengths for robust, high-quality outputs.

Let the probabilities from the ML Decoder and LLM Decoder at time step $t$ be denoted as \(P_{\text{ML}}(w_t \mid \mathbf{z}_i^{\text{new}}, w_{<t})\) and \(P_{\text{LLM}}(w_t \mid \Gamma_i, w_{<t})\), respectively. We adopt a posterior correction strategy inspired by the Product of Experts. The combined probability is:

\begin{align}
    P(w_t) &= \nonumber \\
    &\hspace{-2em} \frac{
        P_{\text{ML}}(w_t \mid \mathbf{z}_i^{\text{new}}, w_{<t})^\lambda \cdot 
        P_{\text{LLM}}(w_t \mid \Gamma_i, w_{<t})^{1-\lambda}
    }{
        \sum\limits_{w \in \mathcal{V}} 
        P_{\text{ML}}(w \mid \mathbf{z}_i^{\text{new}}, w_{<t})^\lambda \cdot 
        P_{\text{LLM}}(w \mid \Gamma_i, w_{<t})^{1-\lambda}
    },
    \label{eq:combined_prob_mixed_input}
\end{align}

where \(w_t\) is the predicted token, \(w_{<t}\) represents the preceding sequence, and \(\lambda \in [0,1]\) controls the balance between two decoders.
This formulation ensures the multiplicative combination of \(P_{\text{ML}}\) and \(P_{\text{LLM}}\), aligning the ML Decoder’s structured precision with the LLM Decoder’s generative flexibility. By adjusting \(\lambda\), we balance deterministic rule-following with creative exploration, enhancing sequence reliability and efficiency.


Our method combines gradient search and symbolic generation through a joint decoder that balances stability from ML and validity from LLM. This integration is simple but effective and has not been used in prior feature transformation methods.

\section{Experiment}

\subsection{Experimental Settings}

\subsubsection{Datasets}
We conducted experiments using datasets from UCIrvine~\cite{uci_dataset_2023}, CPLM~\cite{CPLM_2023}, Kaggle~\cite{howard_kaggle_2023}, and OpenML~\cite{Openml_dataset_2023}. 
The corresponding statistics and tasks are presented in \textbf{Table~\ref{tab:overallresult}}, where 'C' represents classification and 'R' represents regression. 

\subsubsection{Baseline Algorithms}
We compared our method with widely-used feature generation methods, shown in \textbf{Table~\ref{tab:overallresult}}. The details of the baselines are presented in \textbf{Appendix~\ref{app:baseline}}.

\subsubsection{Evaluation Metrics}
We evaluated our framework on both classification and regression tasks. For classification, we use the F1-Score, and for regression, we report 1-RAE (Relative Absolute Error). 
The detailed metrics definitions are provided in \textbf{Appendix~\ref{app:metrics}}.



\subsection{Research Questions}








We aim to address the following research questions:
\textbf{RQ1:} Does the proposed feature transformation framework enhance the downstream performance?
\textbf{RQ2:} What is the impact of the teaming strategy, in terms of error rate, operator ratio, and ablation studies?
\textbf{RQ3:} How well does the proposed framework generalize across different downstream models?
\textbf{RQ4:} How effective are LLMs in feature transformation tasks, and how do they compare to traditional methods?

\subsection{Overall Performance}

To evaluate the effectiveness of our proposed feature transformation framework, we conducted experiments on 23 diverse datasets, covering both classification and regression tasks. 
These datasets vary significantly in size, feature dimensions, and complexity, ensuring a comprehensive assessment of the framework’s generalization ability. 
We compare our Teaming method with several baseline feature transformation approaches, including both traditional and reinforcement learning (RL)-based methods. 
For classification tasks, we use F1-Score as the primary evaluation metric, while for regression tasks, we adopt 1-RAE (inverse relative absolute error). 
The results, summarized in \textbf{Table~\ref{tab:overallresult}}, provide a detailed performance comparison across different datasets.

The ``Original'' results refer to models trained solely on the raw feature set, without any transformed features. The consistent improvements observed across most datasets after applying feature transformation highlight the necessity of generating new, informative features to enhance downstream model performance.

The experimental results reveal that Teaming consistently outperforms other methods across both classification and regression tasks. In classification datasets, Teaming achieves the highest F1-Score, indicating its effectiveness in learning meaningful feature representations. Similarly, in regression datasets, Teaming outperforms other methods in terms of 1-RAE, showcasing its adaptability in different learning tasks.

A key observation is the robust and stable performance of Teaming across diverse datasets. While some baseline methods show performance fluctuations due to varying dataset characteristics, Teaming remains consistently strong regardless of dataset size, feature dimensions, or complexity. This suggests that our framework generalizes well and can be applied effectively in a wide range of tasks.
Furthermore, the results demonstrate that Teaming surpasses both traditional and RL-based methods. Traditional approaches struggle to capture complex data patterns. Even compared to advanced RL-based methods, Teaming still achieves higher performance, highlighting its ability to generate more informative and useful features.
The average ranking at the bottom of \textbf{Table~\ref{tab:overallresult}} further confirms the advantage of Teaming. 

Traditional ML methods (e.g., RDG, LDA, PCA) generally offer fast computation but struggle to capture higher-order or semantically rich transformations, limiting their performance on complex tasks. Pure LLM-based methods (e.g., FeatLLM, CAAFE) exhibit greater expressiveness but are often unstable or biased toward simple operators. Hybrid methods (e.g., MOAT, ELLM-FT, AutoFeat) combine rule-based or optimization strategies with generative components and tend to perform better than single-paradigm approaches. Our proposed Teaming method integrates LLM symbolic reasoning with ML gradient guidance, which results in superior average ranking across all datasets. These results demonstrate both robustness and representational power.

We also check the efficiency in \textbf{Appendix~\ref{app:runtime}}.

\subsection{Teaming Study}

To investigate the impact of different teaming strategies on feature transformation performance, we conducted experiments comparing four different policies. The downstream performance and error rate in \textbf{Table~\ref{tab:ablation_study}} across multiple datasets show how different teaming strategies influence feature transformation quality.

\begin{table}[h]
  \begin{center}
    \caption{Ablation Study. Comparison of Models with Downstream Performance and Error Rate.}
    \label{tab:ablation_study}
    \resizebox{\linewidth}{!}{\begin{tabular}{lcccc}
            \toprule
            \textbf{Dataset} & \makecell{\textbf{Traditional} \\ \textbf{ML}} & \makecell{\textbf{Teaming} \\ \textbf{w/o Search}} & \makecell{\textbf{w/o Decoder} \\ \textbf{Teaming}} & \makecell{\textbf{Teaming} \\ \textbf{Policy}} \\
            \midrule
            Amazon Employee & 93.13\% (0.00\%) & 93.47\% (0.00\%) & 93.45\% (0.00\%) & 93.52\% (0.00\%) \\
            AP-omentum-ovary & 79.64\% (35.00\%) & 79.70\% (25.00\%) & 80.54\% (20.00\%) & 81.39\% (5.00\%) \\
            SpectF & 86.95\% (22.50\%) & 88.02\% (7.73\%) & 89.44\% (2.50\%) & 90.49\% (0.00\%) \\
            German Credit & 72.44\% (85.83\%) & 84.42\% (75.91\%) & 72.75\% (61.25\%) & 85.32\% (70.71\%) \\
            UCI Credit & 80.87\% (17.50\%) & 80.67\% (10.83\%) & 80.80\% (11.36\%) & 80.86\% (0.00\%) \\
            Spam Base & 92.90\% (4.17\%) & 92.92\% (4.09\%) & 93.13\% (1.67\%) & 93.46\% (0.00\%) \\
            Ionosphere & 95.69\% (80.83\%) & 95.74\% (72.27\%) & 95.74\% (77.73\%) & 97.10\% (70.00\%) \\
            Higgs Boson & 69.12\% (54.17\%) & 70.01\% (45.00\%) & 70.36\% (48.33\%) & 70.81\% (37.62\%) \\
            PimaIndian & 80.73\% (54.17\%) & 90.13\% (46.67\%) & 88.99\% (45.00\%) & 91.95\% (21.36\%) \\
            Messidor Feature & 73.02\% (13.18\%) & 74.83\% (5.83\%) & 73.80\% (10.83\%) & 75.61\% (3.75\%) \\
            Wine Quality Red & 62.10\% (20.48\%) & 62.19\% (3.75\%) & 62.35\% (0.83\%) & 62.94\% (0.00\%) \\
            Wine Quality White & 54.52\% (10.83\%) & 54.95\% (8.18\%) & 54.22\% (6.36\%) & 55.18\% (5.83\%) \\
            SVMGuide3 & 81.74\% (32.50\%) & 82.60\% (20.83\%) & 83.84\% (30.71\%) & 84.64\% (20.83\%) \\
            Lymphography & 88.38\% (43.33\%) & 85.28\% (25.00\%) & 88.57\% (15.91\%) & 91.81\% (11.90\%) \\
            \midrule
            Airfoil & 0.5967 (55.00\%) & 0.6311 (48.57\%) & 0.6211 (58.33\%) & 0.6329 (45.00\%) \\
            Housing Boston & 0.4463 (12.50\%) & 0.4482 (2.27\%) & 0.4469 (6.67\%) & 0.4584 (0.00\%) \\
            Openml 586 & 0.6251 (52.50\%) & 0.6405 (33.33\%) & 0.6446 (37.86\%) & 0.6569 (24.17\%) \\
            Openml 589 & 0.5139 (4.17\%) & 0.5937 (0.83\%) & 0.5937 (0.83\%) & 0.5990 (0.00\%) \\
            Openml 607 & 0.6051 (45.00\%) & 0.6181 (41.67\%) & 0.6056 (40.71\%) & 0.6181 (29.17\%) \\
            Openml 616 & 0.4063 (16.36\%) & 0.4066 (15.00\%) & 0.4073 (9.17\%) & 0.4073 (3.33\%) \\
            Openml 618 & 0.4734 (51.67\%) & 0.4823 (45.91\%) & 0.4831 (30.00\%) & 0.4840 (22.50\%) \\
            Openml 620 & 0.5722 (5.91\%) & 0.5748 (1.67\%) & 0.5751 (2.50\%) & 0.5847 (3.33\%) \\
            Openml 637 & 0.2125 (32.73\%) & 0.2588 (38.33\%) & 0.2859 (34.05\%) & 0.3095 (22.50\%) \\
            \bottomrule
        \end{tabular}}
    \end{center}
\end{table}

The \textbf{Traditional ML} setting represents a standard machine learning approach without golden examples. The \textbf{Teaming w/o Search} configuration reduces the search steps to examine the impact of limiting the search space. The \textbf{w/o Decoder Teaming} setting removes the contribution of decoding alignment. Finally, the \textbf{Teaming Policy} applies the full proposed strategy.
The evaluation focuses on two key metrics: downstream performance, which measures the effectiveness of transformed features in enhancing classification and regression tasks, and error rate, which quantifies the proportion of invalid or incorrect sequences generated during feature transformation.

The results in \textbf{Table~\ref{tab:ablation_study}} (where values are presented as ``performance (error rate)'') demonstrate that the Teaming Policy consistently achieves the highest downstream performance while maintaining the lowest error rate across various datasets. One key finding is that reducing search steps negatively impacts feature transformation quality, indicating that a more extensive search process is crucial for generating effective transformations. This underscores the importance of maintaining a well-optimized search space to fully exploit the potential of the transformation framework.

Another critical observation is the role of decoder teaming in enhancing stability. When the decoder teaming mechanism is removed, performance drops significantly, particularly in regression tasks. This suggests that decoder teaming is essential for aligning ML and LLM-generated transformations. Without this alignment, the transformed features may lose consistency, leading to suboptimal results in downstream tasks.  

A particularly notable advantage of the Teaming policy is its ability to reduce error rates. In several datasets, the error rate reaches 0.00\%, demonstrating that the transformed features are highly reliable. 
The ability to consistently generate valid and high-quality feature transformations further reinforces Teaming as a robust and effective approach for improving downstream task performance.

We also studied the importance of two fine-tuning tasks when building the student LLM (\textbf{Appendix~\ref{app:distillation}}), which shows that both tasks enhance the student LLM's capacity and contribute to the final transformation quality.
We also tried different ML-Teacher-Student combinations (\textbf{Appendix~\ref{app:backbone}}) to test the generalization of teaming.

\subsection{Robustness Check}

\begin{figure}[htbp]
    \centering
    \begin{subfigure}[b]{0.22\textwidth}
        \includegraphics[width=\textwidth]{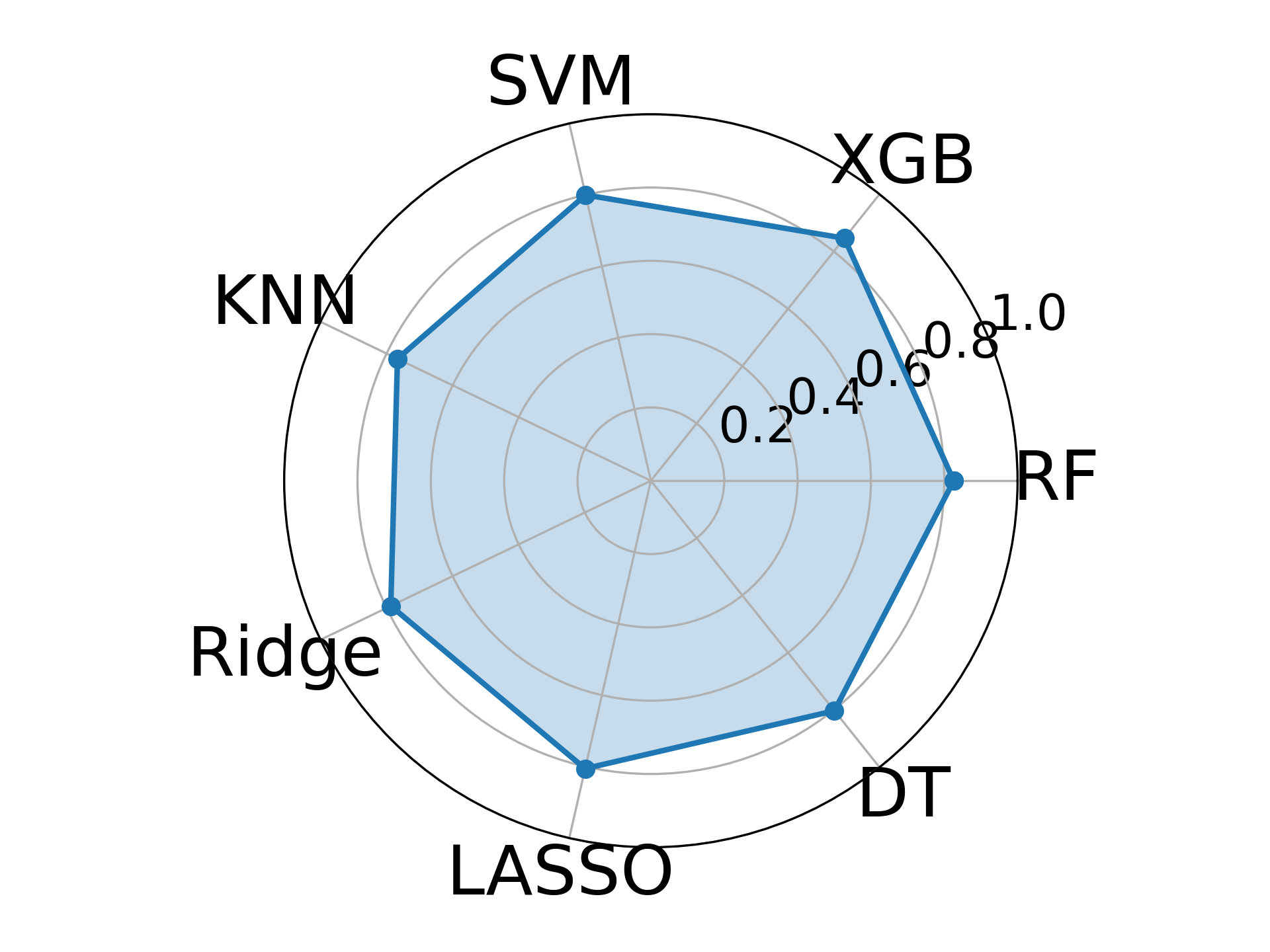} 
        \caption{SVMGuide3 Dataset}
        \label{subfig:robustness_svm}
    \end{subfigure}
    \hfill
    \begin{subfigure}[b]{0.22\textwidth}
        \includegraphics[width=\textwidth]{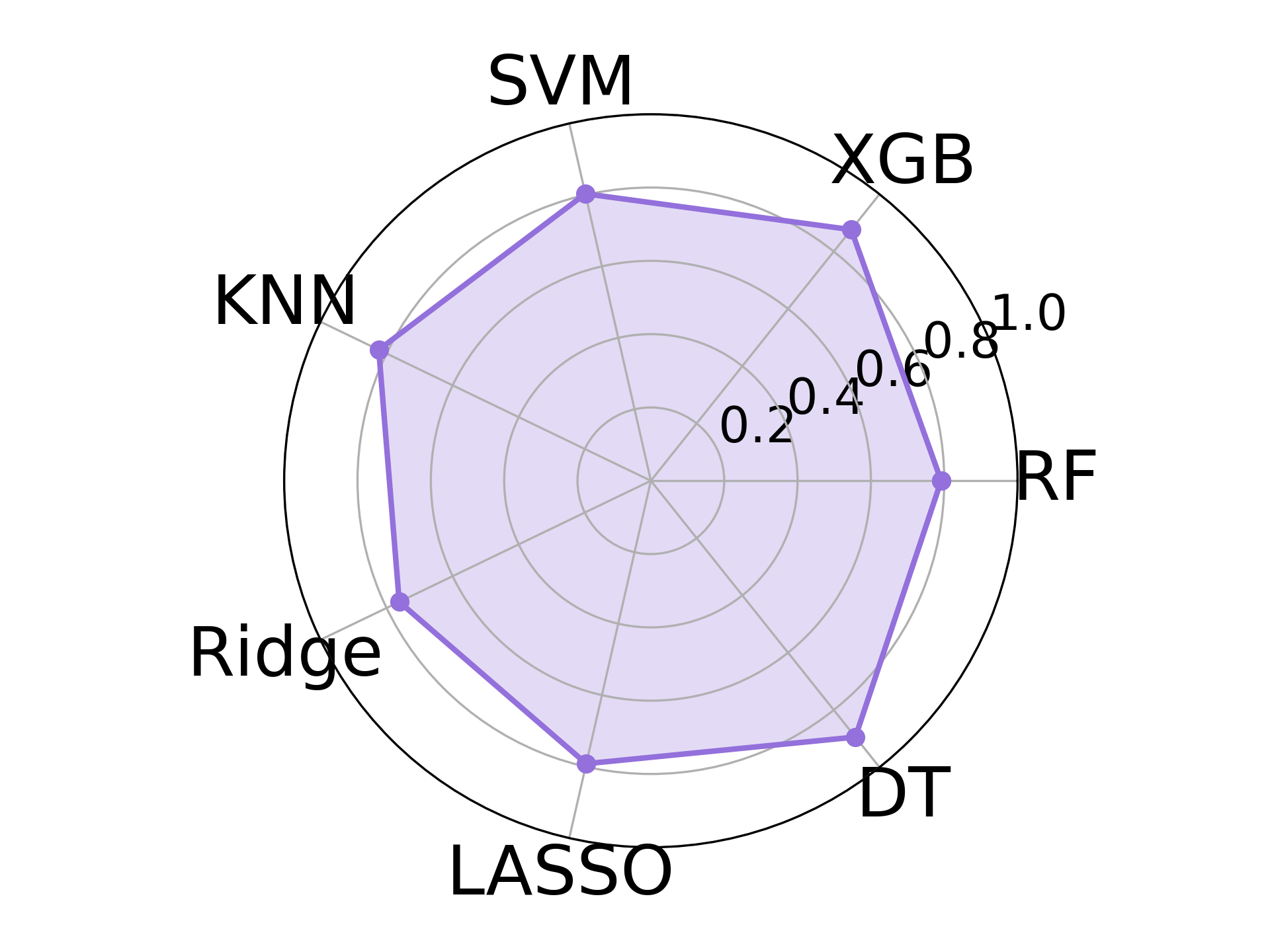} 
        \caption{SpectF Dataset}
        \label{subfig:robustness_spectf}
    \end{subfigure}
    \caption{Robustness Check. The radar charts show the performance across seven downstream models.}
    \label{fig:robustness_check}
\end{figure}

To evaluate the robustness of the teaming policy, we test its performance across multiple different downstream models. Specifically, we apply our feature transformation framework to two datasets: SVMGuide3 and SpectF, utilizing seven different models for downstream tasks.
\textbf{Figure~\ref{fig:robustness_check}} presents the results of the robustness check.

The radar charts show that the transformed features perform consistently across different models, with minimal variation in downstream outcomes. This highlights the robustness and adaptability of our framework across diverse downstream models.

\subsection{LLM for Feature Transformation}
\label{sec:llm_analysis}

We conducted additional experiments to understand how LLMs behave when directly used for feature transformation. 
While these models can generate high-performing features, we also observed some interesting behaviors: they tend to prefer simple operators, show unstable outputs across runs, and naturally focus on important features even without supervision.
These insights help explain both the strengths and limitations of LLM-driven transformations. Full results, visualizations, and analysis are provided in \textbf{Appendix~\ref{app:llmstudy}}.

\section{Related Work}

\noindent\textbf{Feature Transformation.}
Feature transformation, as a key task of data-centric ai~\cite{ying2024feature,ying2024revolutionizing,azim2025biological}, aims to improve the feature space by applying mathematical operations to the original features. Existing methods fall into two main types: (1) \textit{Discrete decision-based methods}, which treat transformation as a discrete search problem~\cite{gong2025agentic,gong2025unsupervised}. Various strategies are adopted to improve search quality, such as heuristic rules~\cite{kanter2015discrete1,khurana2016discrete2,tran2016discrete3,xiao2023traceable,xiao2024evolutionary2}, feature space expansion with selection~\cite{katz2016explorekit}, evolutionary algorithms~\cite{zhu2022evolutionary1,gong2025evolutionary}, and reinforcement learning~\cite{wang2022group, ying2023self,ying2024topology,hu2024reinforcement,azim2024feature}. (2) \textit{Continuous optimization methods}, which embed features into continuous latent spaces and optimize them through gradient-based search~\cite{wang2023reinforcement,ying2024unsupervised,gong2025sculpting,gong2025neuro,baiprivacy}.

\noindent\textbf{LLM for Specific Task.}
Recent studies explore how LLMs assist in feature-related tasks. Aug-imodels~\cite{singh2023augmenting} and Kasneci et al.~\cite{kasneci2024enriching} enrich classical models with LLM-generated embeddings or features. Li et al.~\cite{li2023large} offer a financial-domain review and model selection framework. CAAFE~\cite{hollmann2024large} generates features iteratively based on task context. FeatLLM~\cite{han2024large} applies few-shot prompting to synthesize transformation rules. Kuken et al.~\cite{kuken2024large} analyze LLMs’ preference for simple operators. Jeong et al.~\cite{jeong2024llm} show LLMs can select relevant features using only column names and task descriptions. LFG~\cite{zhang2024dynamic} uses LLM agents and Monte Carlo Tree Search to guide dynamic feature generation. Xu et al.~\cite{xu2024large} integrate LLMs with AutoML to programmatically optimize data pipelines.

\noindent\textbf{ML-LLM Alignment.}
Though ML-LLM alignment is still emerging, several related works offer insights. ARGS~\cite{khanov2024args} adjusts token probabilities during decoding using reward signals to improve output alignment. Kong et al.~\cite{kong2024aligning} model LLMs as discrete-time stochastic systems and apply value function learning via Bellman equations. TreeBoN~\cite{qiu2024treebon} introduces speculative tree search to guide Best-of-N sampling using token-level rewards, balancing efficiency and quality. ELLM-FT~\cite{gong2025evolutionary} adapts evolutionary strategies with few-shot prompting and RL data collection for efficient, high-quality feature transformation.

\section{Conclusion Remarks}
We propose an LLM-ML teaming framework to address the challenges of stability and validity in Generative Feature Transformation.  By combining ML gradient search with LLM symbolic generation, our method produces consistent and high-quality features. 
Experimental results demonstrate that this approach improves transformation reliability and enhances feature expressiveness, achieving a 5\% performance gain. 
This work highlights the promise of LLM-ML collaboration in advancing automated feature engineering.

\section*{Limitations}

While our framework improves feature transformation performance across multiple tasks and models, it still has several limitations.
(1) The student LLM offers a more efficient alternative to the teacher model, but it remains less accurate and more prone to instability during generation.
(2) The framework is task-agnostic and does not incorporate domain-specific information. Incorporating task-aware prompts or fine-tuning may improve relevance and interpretability.
(3) The ML and LLM components are trained independently. A unified or end-to-end training strategy could potentially improve alignment and collaborative performance.
(4) The method has not yet been evaluated in full production pipelines, such as time-series data or enterprise-scale automated systems, where deployment constraints may differ.
(5) The LLM tends to favor simpler operators (e.g., addition, subtraction), which may limit the diversity and complexity of generated transformations in certain tasks.



\bibliography{custom}

\appendix

\section{Important Concept}
\label{app:concept}

\paragraph{Operation Set:} To refine the feature space, we need to apply mathematical operations to existing features to generate new informative features. All operations are collected in an operation set, denoted by $\mathcal{O}$. These operations can be classified as unary and binary operations. The unary operations such as "square", "exp", "log", etc. The binary operations such as "plus", "multiply", "minus", etc.

\begin{figure}[htbp] 
    \centering 
    \includegraphics[width=0.45\textwidth]{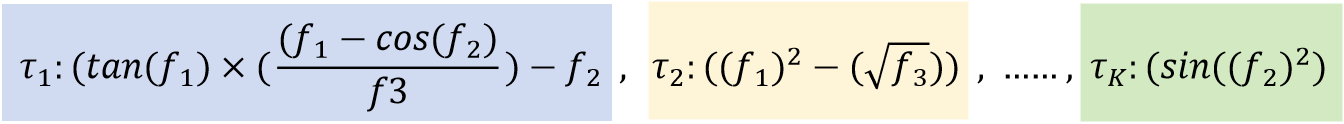} 
    \caption{A Feature Transformation Sequence Example.} 
    \label{fig:app_example_seq} 
\end{figure}

\paragraph{Feature Transformation Sequence:} Assuming a dataset $D = \{X,y\}$ includes the original feature set $X = [f_{1},\cdots,f_{N}]$ and predictive targets $y$. We transform the existing features using mathematical compositions $\tau$ consisting of feature ID tokens and operations to generate new and informative features (\textbf{Figure~\ref{fig:app_example_seq}}). $K$ compositions are adopted to refine $X$ to a better feature space $\tilde{X} = [\tilde{f}_{1},\cdots,\tilde{f}_{K}]$. The collection of the $K$ compositions refers to the feature transformation sequence, which is denoted by $\Gamma = [\tau_{1},\cdots,\tau_{K}]$.

\begin{figure}[htbp]
    \raggedright
    \captionsetup{justification=raggedright,singlelinecheck=false} 
    \begin{subfigure}[b]{0.25\textwidth}
        \includegraphics[width=\textwidth]{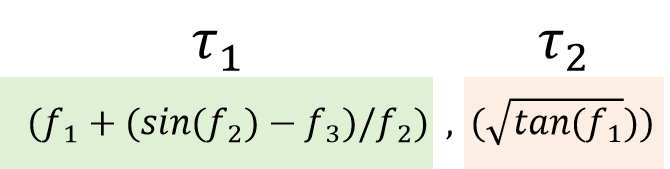}
        \caption{Original Sequence}
        \label{fig:example_postfix_a}
    \end{subfigure}
    \hfill
    \begin{subfigure}[b]{0.49\textwidth}
        \includegraphics[width=\textwidth]{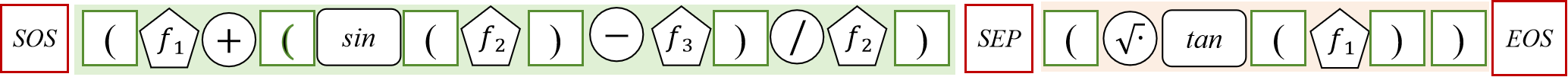}
        \caption{Infix Expression}
        \label{fig:example_postfix_b}
    \end{subfigure}
    \hfill
    \begin{subfigure}[b]{0.33\textwidth}
        \includegraphics[width=\textwidth]{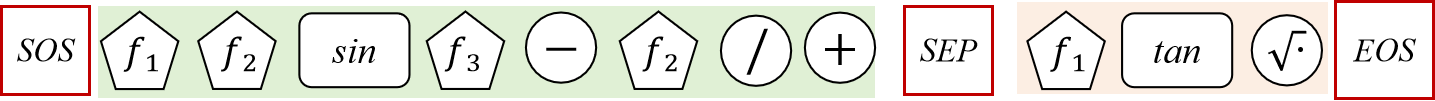}
        \caption{Postfix Expression}
        \label{fig:example_postfix_c}
    \end{subfigure}
    \caption{Different Expressions of Transformation Sequence.}
    \label{fig:example_postfix_all}
\end{figure}

\paragraph{Postfix Expressions:} 
The transformation sequence should be in a computable and machine-learnable format. \textbf{Figure~\ref{fig:example_postfix_a}} shows a transformation sequence with two generated features. The original infix representation (\textbf{Figure~\ref{fig:example_postfix_b}}) has issues like redundancy, semantic sparsity, a high likelihood of illegal transformations, and an overly large search space.

We introduce postfix expressions (\textbf{Figure~\ref{fig:example_postfix_c}}) to solve these problems. 
Postfix expressions don't need many brackets to determine calculation priority. Scanning from left to right suffices to reconstruct the corresponding sequence, greatly reducing sequence-modeling difficulty and computational cost.
They also reduce the ambiguity of the transformation sequence.
Most importantly, it reduces the search space from exponential to a finite set $|C| = |\mathcal{O}| + |X|D + 3$. Here, $|\mathcal{O}|$ represents the operation set size, $|X|$ is the original feature set dimension, $D$ is feature numbers, and $3$ refers to start tokens $<SOS>$, separation token $<SEP>$, and end token $<EOS>$.

\section{Feature Transformation Prompt}
\label{app:prompt}

\begin{figure*}[t]
    \centering
    \includegraphics[width=0.99\textwidth]{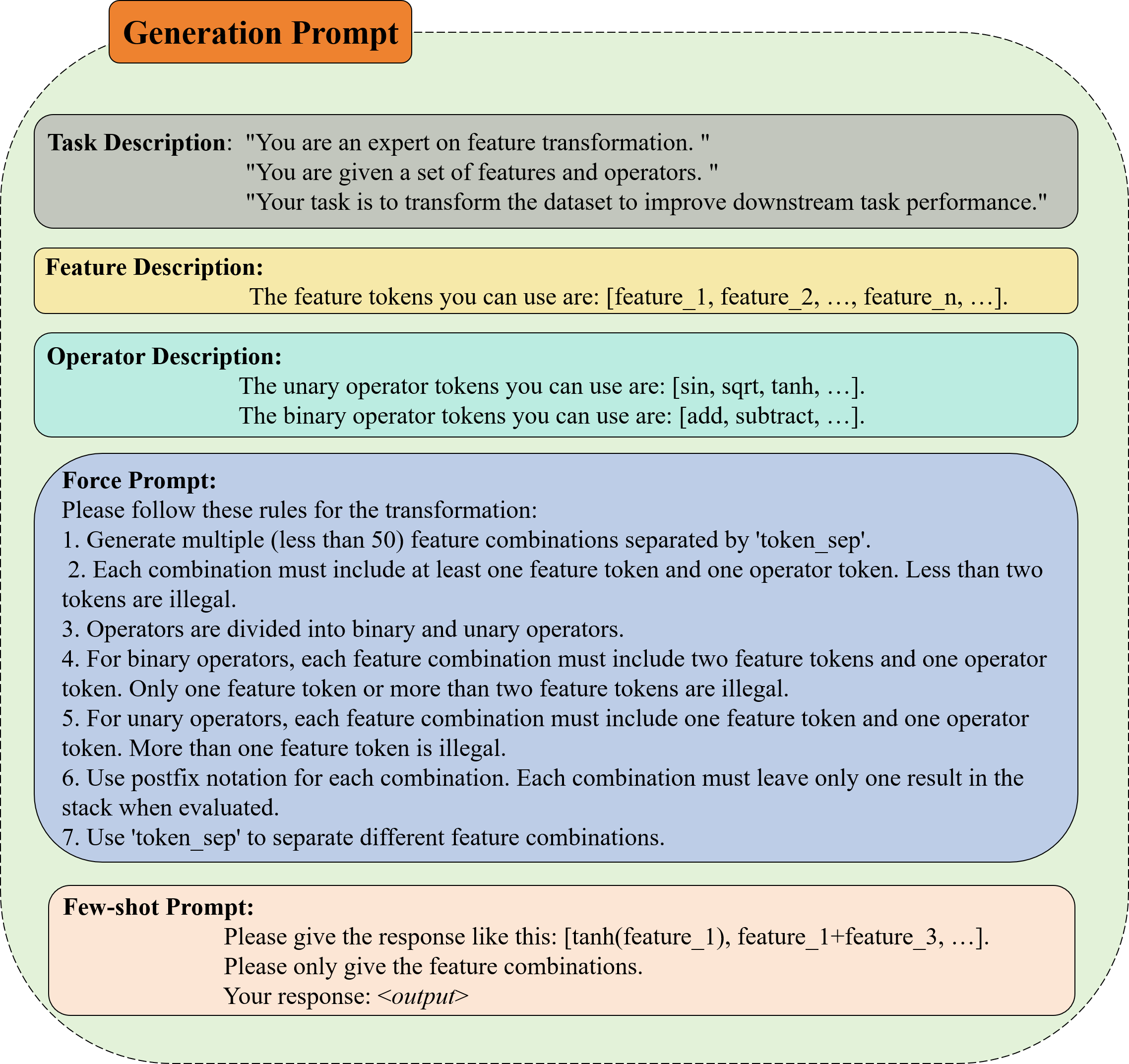} 
    \caption{The prompt details.}
    \label{fig:prompt}
\end{figure*}

This is a detailed description of the generation prompt used in the study. The prompt is designed to guide LLMs on feature transformation to improve downstream task performance.

The prompt is structured into several parts. The \textbf{Task Description} section introduces the role of the expert and the overall goal of the transformation. It states that the expert is given a set of features and operators and is tasked with dataset transformation.
The \textbf{Feature Description} part lists the available feature tokens that can be used in the transformation.
The \textbf{Operator Description} section details the unary and binary operator tokens available. The unary operator tokens include [sin, sqrt, tanh, ...], and the binary operator tokens include [add, subtract, ...].
The \textbf{Force Prompt} section enforces several rules for the transformation. These rules ensure that the generated feature combinations are valid and follow the specified format. For example, it requires generating multiple (less than 50) feature combinations separated by 'token\_sep', each combination to include at least one feature and one operator token, and also has specific rules for binary and unary operators in terms of the number of feature tokens they can operate on. It also mandates the use of postfix notation and the use of 'token\_sep' to separate different combinations.
Finally, the \textbf{Few-shot Prompt} section provides an example of how the response should be formatted and requests that only the feature combinations be given in the response.

\section{Student LLM Feature Transformation}
\label{app:distillation}

To construct a reliable student LLM, we employ two fine-tuning tasks: (1) \textbf{Sequence Generation Task} and (2) \textbf{Performance Prediction Task}.

\subsection{Sequence Generation Task}
The student LLM is fine-tuned to generate feature transformation sequences, leveraging patterns and principles captured in the teacher LLM's golden example database. The input prompts provided to the student LLM are consistent with those used for the teacher LLM in the golden example generation phase.
To optimize this task, a cross-entropy loss function is employed: $\mathcal{L}_{\text{seq}} = -\sum_{n=1}^{m} \log(P_{\text{LLM}}(\Gamma_n))$,
where $\Gamma_n$ represents the $n$-th golden example, $P_{\text{LLM}}(\Gamma_n)$ is the student LLM generating probability, and $m$ is the total number of sequences.
This fine-tuning process distills the teacher LLM’s knowledge into the student LLM, enabling it to explore complex and innovative feature transformations while adhering to the postfix expression format. This ensures low error rates during decoding.

\subsection{Performance Prediction Task}

The student LLM is also trained to predict the effectiveness of its generated feature transformation sequences $\Gamma$. Let $v(\Gamma)$ denote the actual performance of a sequence $\Gamma$, and $\hat{v}(\Gamma)$ is predicted by the LLM. The MSE loss is defined as: $\mathcal{L}_{\text{perf}} = \frac{1}{m} \sum_{i=1}^{m} \left( \hat{v}(\Gamma_i) - v(\Gamma_i) \right)^2$,
with $m$ training samples.
By learning to associate transformation patterns with performance metrics, the student LLM prioritizes high-quality transformations while discarding suboptimal ones. This dual-task training enhances the overall efficiency and effectiveness of the feature transformation framework.

\subsection{Distillation Study}


We conducted experiments on those two fine-tuning tasks to illustrate their importance.

\begin{table}[htbp]
    \centering
    \caption{Distillation Study on OpenML 586 Dataset.}
    \resizebox{\linewidth}{!}{
    \begin{tabular}{lcc}
        \toprule
        Dataset & Performance $\uparrow$ & Error Rate $\downarrow$ \\
        \midrule
        with Performance Prediction & 0.6569 & 24.17\%  \\
        w/o Performance Prediction & 0.6519 & 34.00\%  \\
        \bottomrule
    \end{tabular}
    }
    \label{tab:distill}
\end{table}

\textbf{Table~\ref{tab:distill}} presents a comparative study on the effect of performance prediction. The results indicate that omitting the performance prediction task leads to a decrease in downstream performance and an increase in error rate. This highlights the importance of performance prediction in improving the model’s effectiveness.

\section{Baselines}
\label{app:baseline}
Here are the baselines that we used in our experiments.
\begin{itemize} [nosep]
    \item \textbf{RDG:} Generates feature-operation-feature transformation records at random to create a new feature space.
    \item \textbf{LDA~\cite{blei2003latent}:} A matrix factorization-based method to obtain the factorized hidden state as the generated feature space.
    \item \textbf{ERG:} Applies operations on each feature to expand the feature space and selects crucial features as new features.
    \item \textbf{NFS~\cite{chen2019neural}:} Models the transformation sequence of each feature and uses reinforcement learning (RL) to optimize the entire feature generation process.
    \item \textbf{AFAT~\cite{horn2020autofeat}:} An enhanced version of ERG that repeatedly generates new features and uses multi-step feature selection to select informative ones.
    \item \textbf{PCA~\cite{mackiewicz1993principal}} Generates new features through linear feature correlation.
    \item \textbf{TTG~\cite{khurana2018feature}:} Formulates the transformation process as a graph and implements an RL-based search method to find the best feature set.
    \item \textbf{GRFG~\cite{wang2022group}:} Utilizes three collaborative reinforced agents to conduct feature generation and proposes a feature grouping strategy to accelerate agent learning.
    \item \textbf{MOAT~\cite{wang2023reinforcement}:} Utilizes a search-based method for better feature space representation, leading to better decoding operator sequences.
    \item \textbf{FeatLLM~\cite{han2024large}:} A recent approach that leverages large language models for few-shot symbolic feature engineering, enabling interpretable transformations with minimal supervision.
    \item \textbf{CAAFE~\cite{hollmann2023large}:} A context-aware automated feature engineering framework that uses LLMs to iteratively refine and select transformations based on dataset metadata and task descriptions.
    \item \textbf{AutoFeat~\cite{horn2019autofeat}:} A classic Python library for automatic feature engineering, generating polynomial and interaction features followed by selection based on statistical relevance.
    \item \textbf{OpenFE~\cite{zhang2023openfe}:} An open-source framework that applies model-agnostic, gradient-guided search to select effective feature transformations.
    \item \textbf{ELLM-FT~\cite{gong2025evolutionary}:}A hybrid evolutionary learning method where LLMs generate transformation candidates and are filtered using reinforcement-style utility scoring.
\end{itemize}

\section{Evaluation Metrics}
\label{app:metrics}

For classification tasks, we use the F1-Score as the evaluation metric:
\begin{equation}
    F1 = 2 \cdot \frac{\text{Precision} \cdot \text{Recall}}{\text{Precision} + \text{Recall}}
\end{equation}
where Precision = $\frac{TP}{TP+FP}$ and Recall = $\frac{TP}{TP+FN}$.

For regression tasks, we report 1-RAE (Relative Absolute Error):
\begin{equation}
    1\text{-RAE} = 1 - \frac{\|\bm{y}_{pred} - \bm{y}_{real}\|_1}{\|\bm{y}_{real} - \bar{\bm{y}}_{real}\|_1}
\end{equation}
where $\bm{y}_{pred}$ is the predicted value, $\bm{y}_{real}$ is the true value, and $\bar{\bm{y}}_{real}$ is the mean of the true values.

\section{Configurations}
\label{app:configurations}

All experiments were conducted on the Ubuntu 22.04.3 LTS operating system, with a 13th-generation Intel(R) Core(TM) i9-13900KF CPU and an NVIDIA GeForce RTX 4090 GPU. The experiments were implemented using Python 3.11.5 and PyTorch 2.0.1.

\section{Efficiency Study}

\label{app:runtime}

The goal of this experiment is to evaluate whether our framework can achieve considerable results with fewer search iterations, thereby improving the efficiency of the feature transformation process. Specifically, we compare the performance of the Teaming policy and the ML-Based policy under different search rounds, measuring downstream task performance as a function of the number of search iterations. This comparison allows us to assess how effectively the Teaming strategy optimizes feature transformations in the latent space.

\begin{table}[htbp] 
    \centering 
    \caption{Efficiency Check Results} 
    \label{tab:efficiency_check} 
    \resizebox{\linewidth}{!}{
    \begin{tabular}{lcccc}
        \toprule
        \multirow{2}{*}{Dataset} & \multicolumn{2}{c}{Teaming Policy} & \multicolumn{2}{c}{ML-Based Policy} \\
        \cline{2 - 5}
        & epoch & second/epoch & epoch & second/epoch \\
        \midrule
        Openml 586 & 8 & 2.14 & 22 & 1.67 \\
        \bottomrule
    \end{tabular}
    }
\end{table}


The results in Table~\ref{tab:efficiency_check} indicate that while the Teaming policy requires slightly more time per epoch (2.14 seconds) compared to the ML-Based policy (1.67 seconds), it converges significantly faster, requiring only 8 epochs, whereas the ML-Based policy takes 22 epochs to reach convergence.
This suggests that the Teaming strategy accelerates the feature transformation process by guiding the search more effectively, reducing the total number of iterations required to reach an optimal transformation. 
Despite the per-epoch time being approximately 28.1\% longer than the ML-Based policy, the total computation time for convergence is 17.12 seconds for the Teaming policy (8 × 2.14), compared to 36.74 seconds for the ML-Based policy (22 × 1.67). This represents an overall 53.4\% reduction in total computation time.

To complement this analysis, we also benchmark the end-to-end runtime of our method against other ML-based and LLM-based feature engineering methods. \textbf{Table~\ref{tab:runtime_comparison}} summarizes the average runtime (in seconds) for each method to complete the transformation pipeline on the OpenML 586 dataset.

\begin{table}[htbp]
    \centering
    \caption{End-to-End Runtime Comparison Across Methods}
    \label{tab:runtime_comparison}
    \resizebox{\linewidth}{!}{
    \begin{tabular}{lccccccc}
        \toprule
        \textbf{Method} & CAAFE & OpenFE & AutoFeat & MOAT & Pure LLM & Teaming \\
        \midrule
        Runtime (s) & 94.69 & 6.55 & 37.83 & 36.74 & 8.32 & 17.12 \\
        \bottomrule
    \end{tabular}
    }
\end{table}

As shown, our method runs faster than heavy pipelines such as CAAFE and FSNS, while maintaining competitive efficiency with Pure LLM-based generation. Despite being slower than OpenFE, which applies simple transformations, our approach provides a more robust balance between runtime and transformation quality.

\section{backbones Study}
\label{app:backbone}

To evaluate the generality of our LLM-ML teaming framework, we conducted an experiment using different combinations of teacher and student LLMs. Specifically, we tested three student LLMs: LLaMA-3, GPT-2, and BART. We also thried a diverse set of teacher LLMs, including GPT-4o, o3-mini, o1-mini, LLaMA 3.2-405B, LLaMA 4, Claude 3, and DeepSeek V3. 
Under Llama-3 student LLM, we tried different ML methods, including LSTM and Transformer decoders.
Each configuration was integrated into our teaming framework, and the downstream performance was measured on a representative regression task from the OpenML 586 benchmark. \textbf{Table~\ref{tab:llmcompare}} reports the average prediction accuracy, along with the corresponding error rate in parentheses.

\begin{table*}[t]
\centering
\caption{Performance and error rate of different student-teacher LLM combinations on OpenML-586 dataset.}
\label{tab:llmcompare}
\resizebox{\textwidth}{!}{
\begin{tabular}{lccccccc}
\toprule
\textbf{Student $\backslash$ Teacher} & \textbf{GPT-4o} & \textbf{o3-mini} & \textbf{o1-mini} & \textbf{LLaMA 3.2-405B} & \textbf{LLaMA 4} & \textbf{Claude 3} & \textbf{DeepSeek V3} \\
\midrule
\textbf{LLaMA-3 (LSTM)} & 0.6569 (24.17\%) & 0.6807 (23.33\%) & 0.6728 (29.17\%) & 0.6807 (23.33\%) & 0.6807 (23.33\%) & 0.6807 (25.00\%) & 0.6688 (19.17\%) \\
\textbf{LLaMA-3 (Transformer)} & 0.6555 (17.45\%) & 0.6555 (17.45\%) & 0.6555 (17.45\%) & 0.6538 (23.33\%) & 0.6555 (17.45\%) & 0.6555 (17.45\%) & 0.6555 (17.45\%) \\
\textbf{GPT-2 (LSTM)} & 0.6446 (39.80\%) & 0.6669 (38.33\%) & 0.6677 (47.73\%) & 0.6694 (34.09\%) & 0.6694 (47.73\%) & 0.6669 (38.33\%) & 0.6336 (25.51\%) \\
\textbf{BART (LSTM)} & 0.6409 (33.33\%) & 0.6723 (39.55\%) & 0.6683 (49.09\%) & 0.6728 (29.17\%) & 0.6683 (49.09\%) & 0.6728 (29.17\%) & 0.6336 (25.51\%) \\
\bottomrule
\end{tabular}
}
\end{table*}

The results demonstrate that our framework generalizes well across different architectures. Accuracy remains consistent across most teacher models, with variation typically within 2–3\%. Student models based on modern LLMs, such as LLaMA-3, achieve the best overall performance, while older architectures like GPT-2 and BART yield slightly lower accuracy and higher error rates. Interestingly, the LSTM-based decoder for LLaMA-3 outperforms the Transformer-based version in several settings, suggesting that sequential decoding may be more effective for symbolic generation tasks. Furthermore, combinations involving GPT-4o and Claude 3 as teacher models consistently deliver strong performance, highlighting the compatibility of our framework with both proprietary and open-source LLM ecosystems.

These findings confirm that the teaming strategy is robust and architecture-agnostic, making it a practical choice for real-world applications that may involve heterogeneous LLM backbones.

\section{LLM for Feature Transformation}
\label{app:llmstudy}
This section investigates the use of LLMs for direct feature transformation tasks, exploring their strengths, inherent limitations, and unexpected findings.

\subsection{Transformation Performance}
We evaluated the performance of the teacher LLM (GPT-4o) and the student LLM (Llama 3.2-3B) in generating feature transformation sequences directly from prompts. \textbf{Table~\ref{tab:llmvsml}} shows the result comparison on the OpenML 586 dataset.

\begin{table}[htbp]
    \centering
    \caption{LLM V.S. ML on OpenML 586 Dataset.}
    \resizebox{\linewidth}{!}{
    \begin{tabular}{lccc}
        \toprule
        Metric & Teacher LLM & Student LLM & ML \\
        \midrule
        Performance & 0.7196 & 0.6867 & 0.6251 \\
        Error Rate & 1.53\% & 20.34\% & 52.50\% \\
        Cost (Dallor) & 1.93 & 0 & 0 \\
        Interpretability & \textcolor{red}{\xmark} & \textcolor{red}{\xmark} & \textcolor{green}{\cmark} \\ 
        \bottomrule
    \end{tabular}
    }
    \label{tab:llmvsml}
\end{table}

The teacher LLM achieved the highest performance, significantly outperforming both the student LLM and traditional ML methods. This highlights the superior ability of LLMs to identify meaningful feature transformations that enhance model effectiveness.  

A key advantage of the teacher LLM is its capacity to generate diverse and high-quality transformations, often uncovering patterns that traditional ML methods might overlook. However, its lack of interpretability remains a notable limitation.  

The student LLM, distilled from data generated by the teacher LLM, maintains a performance level close to that of the teacher while exhibiting a higher error rate. This suggests that knowledge distillation to a smaller model introduces some degradation in feature transformation accuracy. Nevertheless, the student model offers a cost-effective alternative, as it incurs no additional computational expenses compared to the teacher LLM. Moreover, the probability distribution of each transformation step enables the implementation of the decoder teaming policy.  

Conversely, while traditional ML approaches exhibit lower performance, they remain highly interpretable and computationally efficient, making them a viable option in scenarios where explainability is a priority.


\subsection{Stability}

Powerful LLMs like GPT-4o can generate feature crosses with better downstream performance, but their black-box nature and lack of stability make cost control challenging. To achieve more diverse outputs, we lower the temperature when generating sequences. However, a lower temperature also increases randomness, making it unclear when to stop.

\begin{figure}[htbp]
    \centering
    \begin{subfigure}[b]{0.15\textwidth}
        \includegraphics[width=\textwidth]{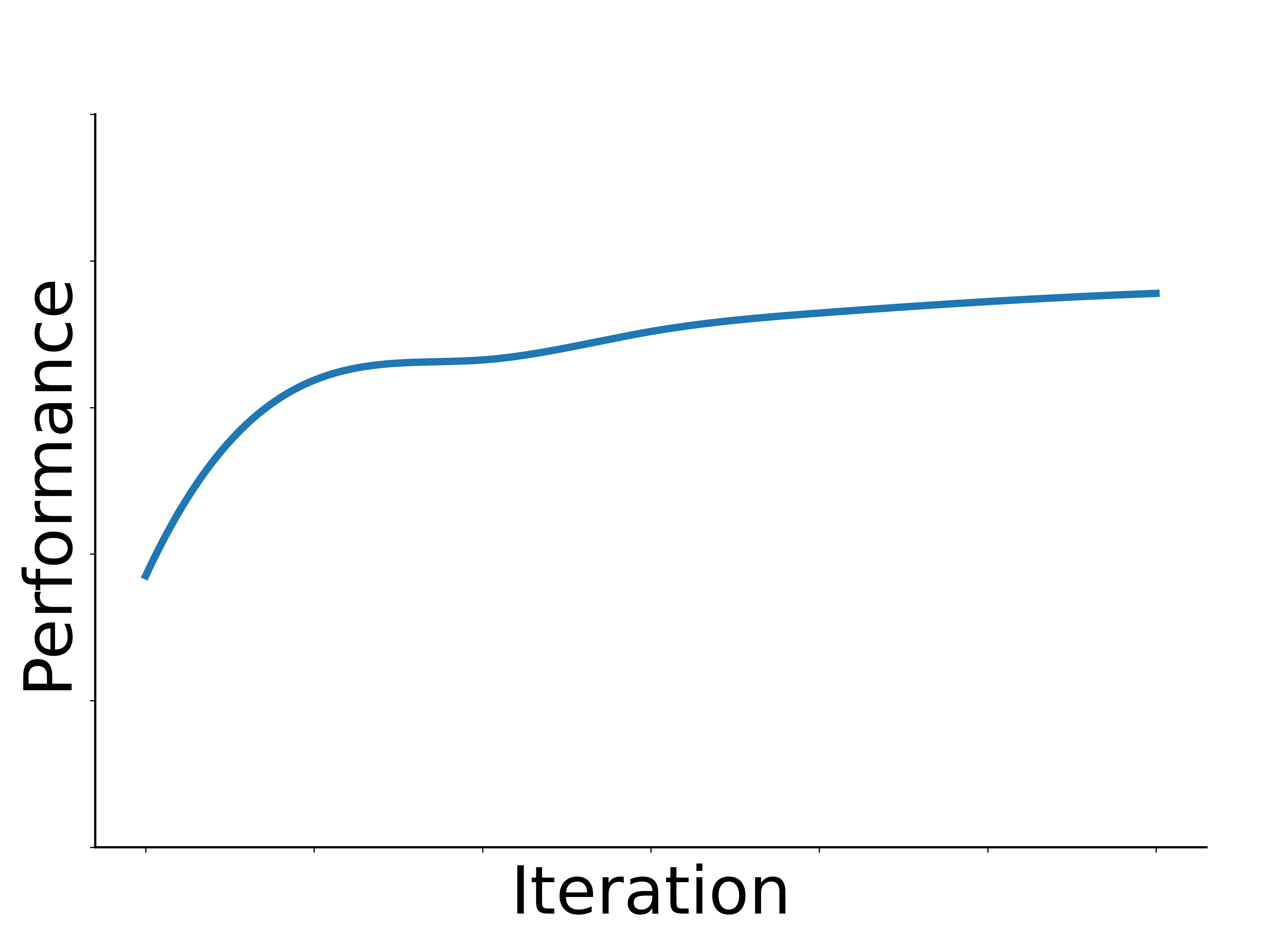} 
        \caption{Traditional ML}
        \label{subfig:stability_ml}
    \end{subfigure}
    \begin{subfigure}[b]{0.30\textwidth}
        \includegraphics[width=\textwidth]{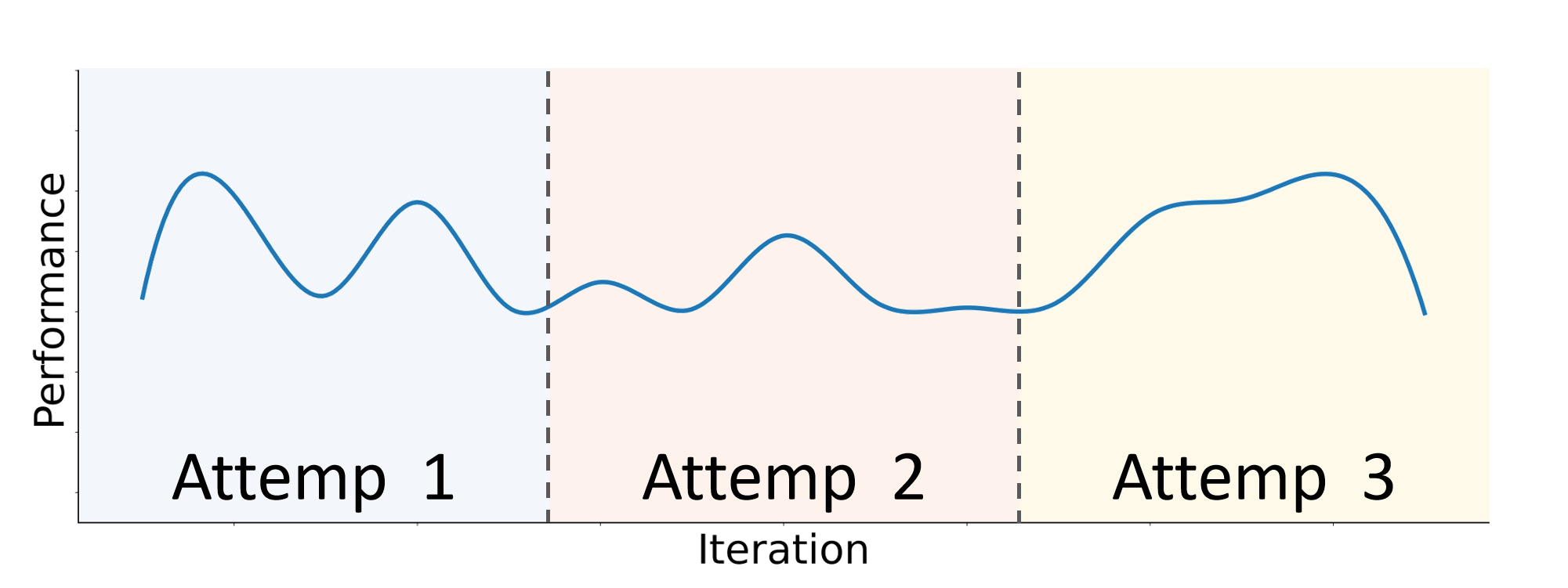} 
        \caption{LLM}
        \label{subfig:stability_gpt}
    \end{subfigure}
    \caption{Stability Comparison. Traditional ML ensures steady improvements, while LLM fluctuates unpredictably.}
    \label{fig:stability}
\end{figure}

\textbf{Figure~\ref{fig:stability}} compares two approaches. 
The traditional ML method with a search policy ensures stable performance improvements (\textbf{Figure~\ref{subfig:stability_ml}}). 
However, the LLM-based generation is unpredictable.
There is always the possibility that trying 1,000 more times might yield significantly better feature crosses, but we cannot afford endless trials. In \textbf{Figure~\ref{subfig:stability_gpt}}, each attempt (segmented regions) exhibits significant fluctuations, lacking a consistent upward trend. Some attempts yield improved results, while others regress, making it uncertain whether further trials will enhance performance or introduce redundancy. If the temperature is too high, we risk redundancy and wasted cost; if it is too low, we never know if the next attempt will be better or worse, making every decision a gamble. 
This dilemma makes temperature tuning challenging, as it creates uncertainty in balancing efficiency and diversity.

\subsection{Operator Ratio}

This experiment investigates whether LLMs exhibit a tendency to prefer simple operators during feature transformation tasks, as suggested by prior research~\cite{kuken2024large}. By analyzing the frequency of operator usage in transformation sequences, we observed the following trends:

\begin{figure}[htbp]
    \centering
    \begin{subfigure}[b]{0.15\textwidth}
        \includegraphics[width=\textwidth]{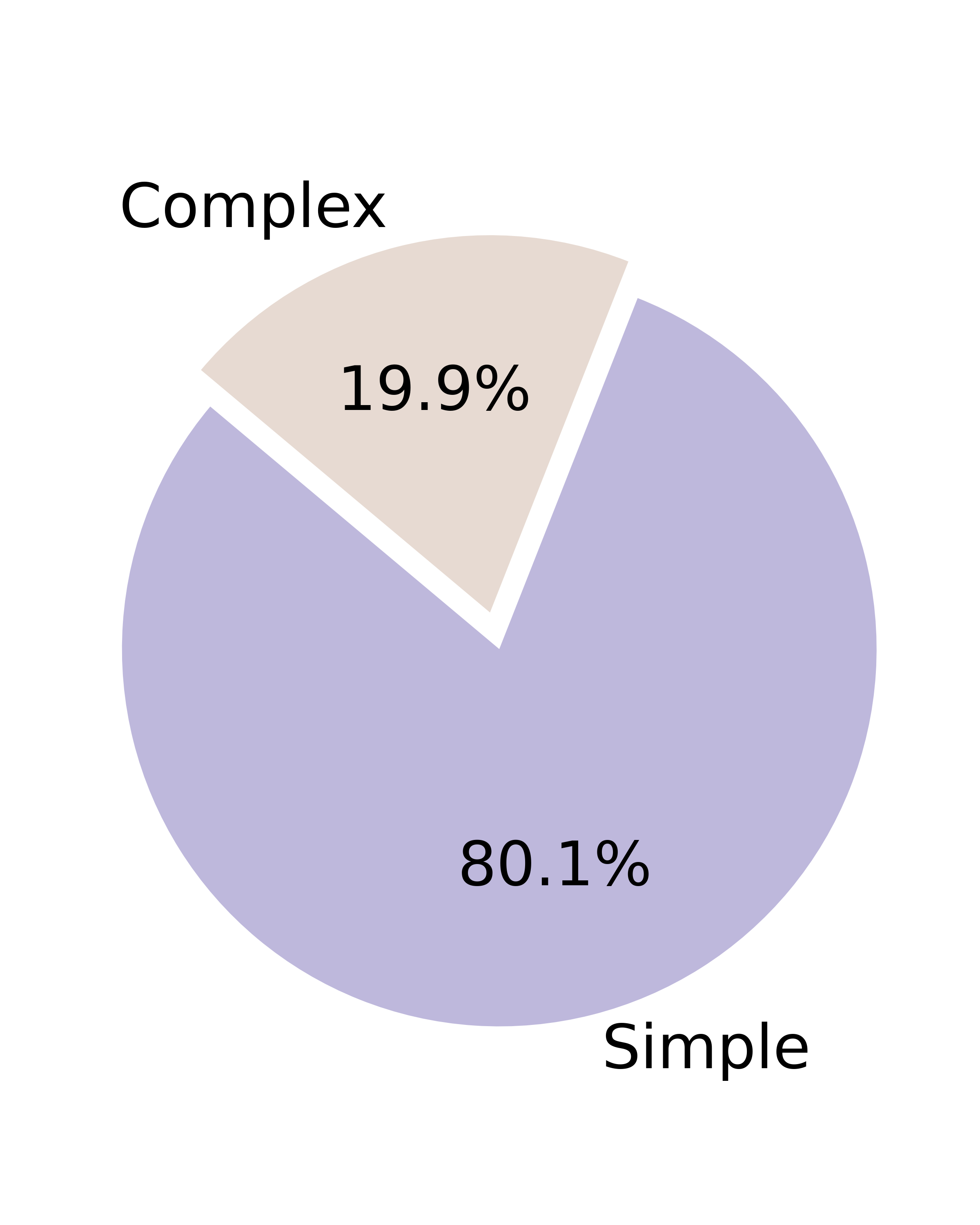} 
        \caption{Teacher LLM}
        \label{subfig:apioperator}
    \end{subfigure}
    \begin{subfigure}[b]{0.15\textwidth}
        \includegraphics[width=\textwidth]{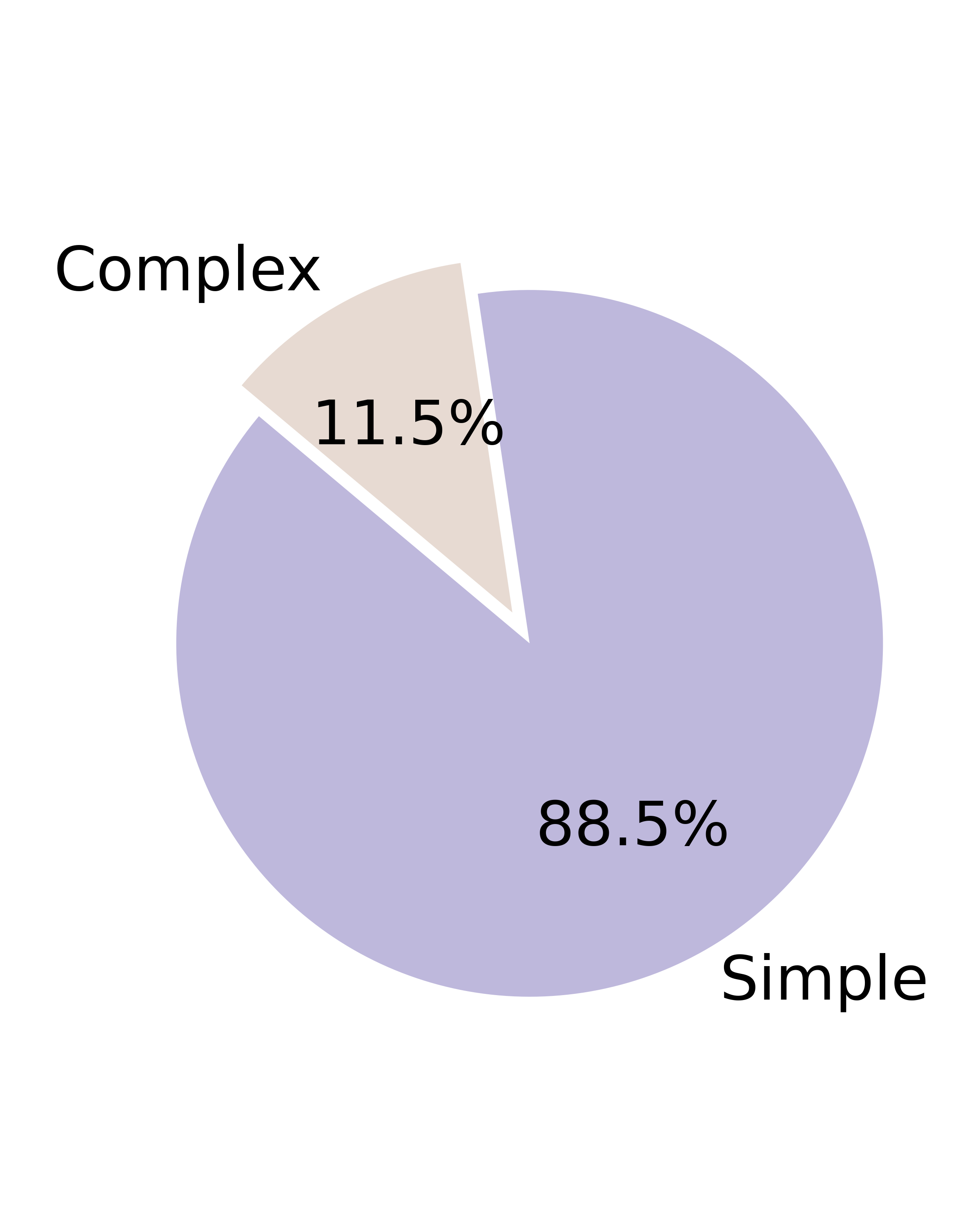} 
        \caption{Student LLM}
        \label{subfig:tunableoperator}
    \end{subfigure}
    \begin{subfigure}[b]{0.15\textwidth}
        \includegraphics[width=\textwidth]{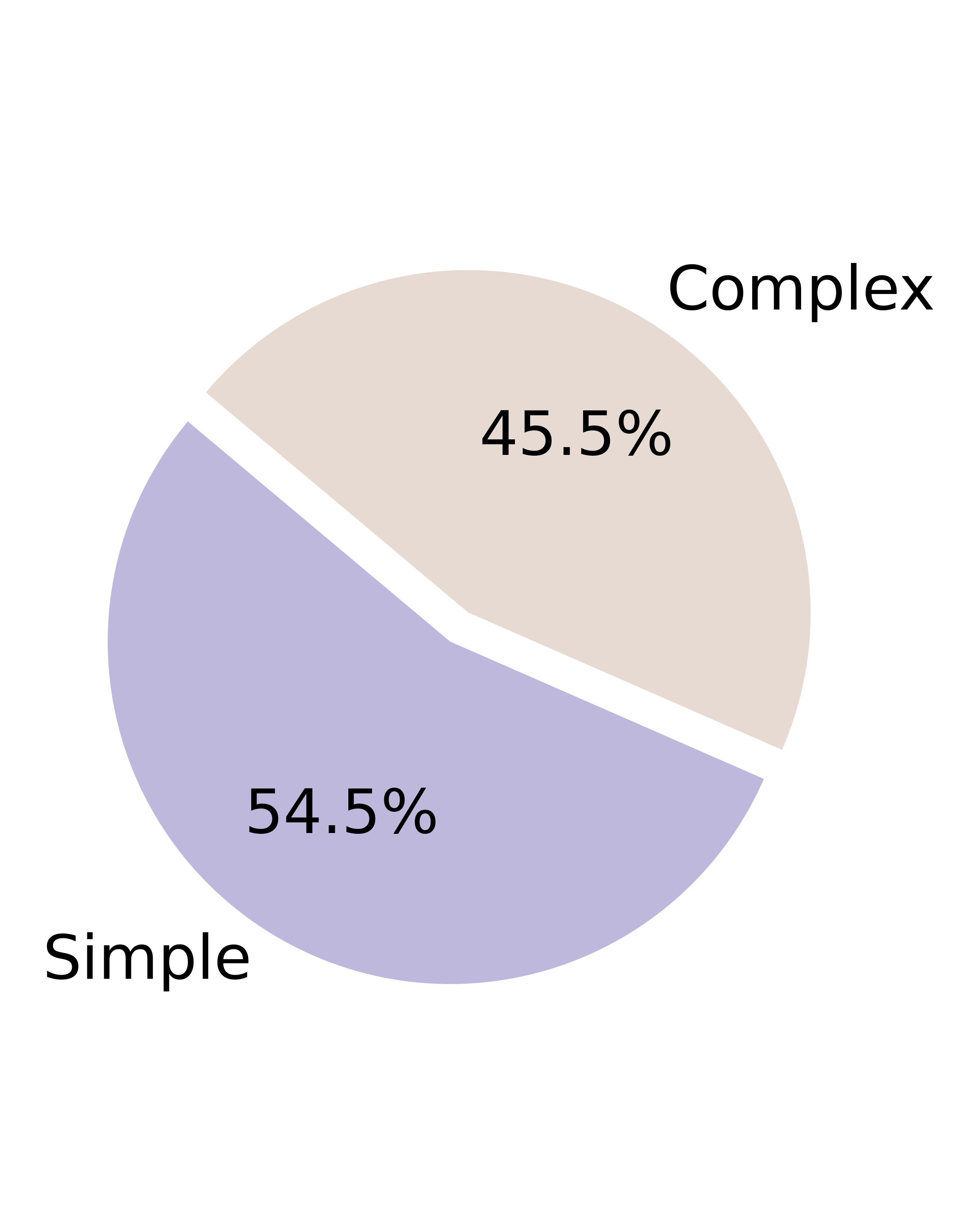} 
        \caption{ML}
        \label{subfig:mloperator}
    \end{subfigure}
    \caption{Operator Usage. They illustrate the ratio of simple vs. complex operators used by different methods.}
    \label{fig:operatorusage}
\end{figure}

As illustrated in \textbf{Figure~\ref{subfig:apioperator}}, the teacher LLM mostly relies on simple operators, with complex operators accounting for only one-fifth of its transformations. The student LLM (\textbf{Figure~\ref{subfig:tunableoperator}}) exhibits an even stronger preference for simple operators, further reducing the use of complex transformations. In contrast, traditional ML methods (\textbf{Figure~\ref{subfig:mloperator}}) demonstrate a more balanced distribution between simple and complex operators, with an approximately equal proportion of each.

These results indicate that LLMs exhibit a clear preference toward simple operations (e.g., addition and subtraction) while under-utilizing more advanced transformations, such as logarithmic and exponential functions. This preference may restrict the diversity and effectiveness of generated features, particularly for datasets that benefit from complex mathematical transformations. These findings corroborate the prior study~\cite{kuken2024large} and highlight the need for strategies to encourage using complex operators (e.g., prompt engineering or fine-tuning). 

We tried different prompts to encourage LLMs to use more complex operators.
To encourage LLM to use more complex operators, we designed some prompts and experimented. Specifically, we:
\begin{itemize} [nosep]
    \item Modified prompts to explicitly encourage diversity in operator use, increasing complex operator rates from 19.9\% to 37.6\%;
    \item Added chain-of-thought reasoning examples, which reduced repetition and encouraged compositionality;
    \item Applied rule-based post-filtering to discard overly simple expressions, improving operator complexity without degrading performance.
\end{itemize}
These techniques demonstrate the controllability of symbolic output generation in LLMs.
However, LLMs prefer simple operators rather than complex operators. The reason could be that they don't want to make a mistake. After all, the first priority of LLMs is to answer the question rather than give the correct answer. That's also the reasons why LLMs may create some unreliable answers, also known as LLM hallucination.

\ul{Does it hurt the overall performance?} There is no empirical evidence suggesting that the use of more simple operators negatively impacts downstream performance. Feature transformation is inherently an open-ended problem: there is no single optimal solution, but rather multiple valid paths to effective representations. Much like the notion that “there is no absolute ranking in art,” the diversity of transformation strategies reflects the creative space of this task. This open-ended nature further bridges feature transformation with natural language generation, making it particularly well-suited for solutions based on large language models.

\subsection{Findings of Feature Selection}

To assess whether the LLMs truly understand the dataset and task—an essential factor for the validity of previous results—we analyze the distribution of feature usage in the generated transformation sequences.

\begin{figure}[htbp]
    \centering
    \begin{subfigure}[b]{0.22\textwidth}
        \includegraphics[width=\textwidth]{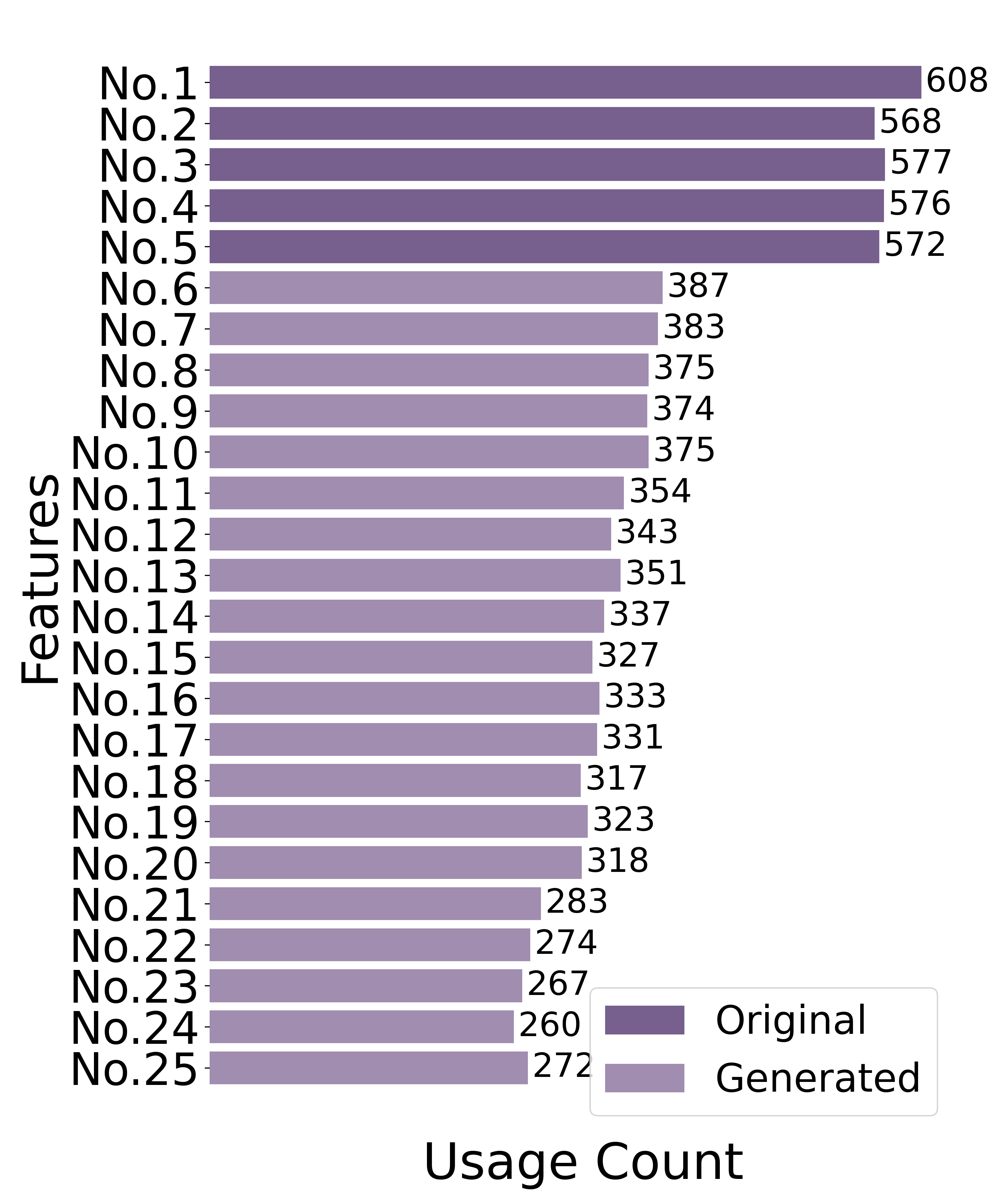}
        \caption{Teacher (GPT-4o).}
        \label{subfig:feature_frequency_API}
    \end{subfigure}
    \begin{subfigure}[b]{0.22\textwidth}
        \includegraphics[width=\textwidth]{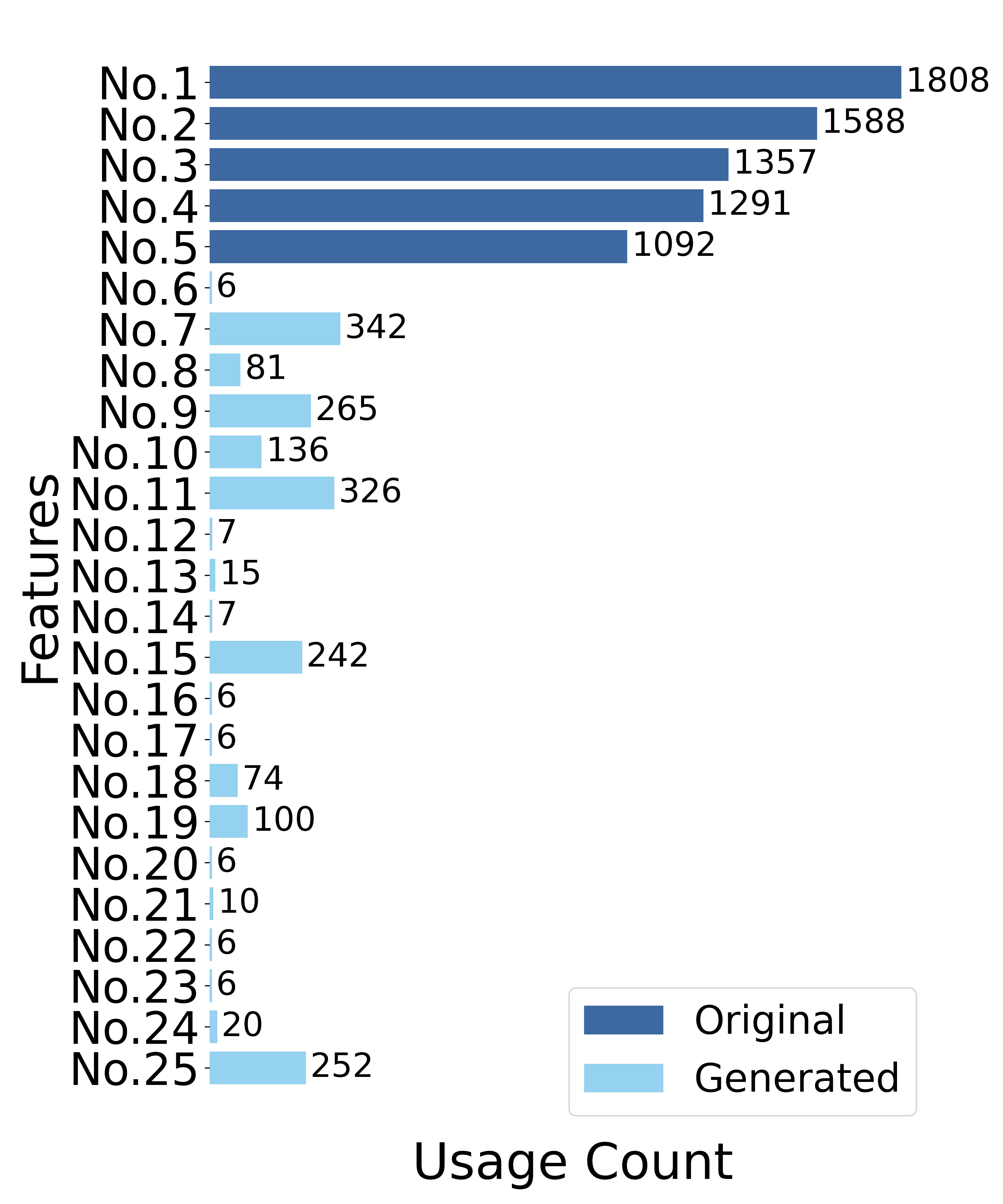}
        \caption{Student (Llama 3.2).}
        \label{subfig:feature_frequency_Llama}
    \end{subfigure}
    \caption{Feature Usage Distribution. The bar charts depict how frequently each feature appears in transformation sequences by different LLMs.}
    \label{fig:feature_frequency}
\end{figure}

In the LLM prompt setup (\textbf{Figure~\ref{fig:prompt}}), both features and operators are treated as tokens, making it crucial to determine whether the LLMs recognize their actual significance. We select an OpenML dataset. The first five features are original, while the next twenty are generated from them. We found strong feature selection characteristics, as shown in \textbf{Figure~\ref{fig:feature_frequency}}.

\textbf{Figure~\ref{subfig:feature_frequency_API}} shows that GPT-4o has a clear preference for the first five original features, with a steep drop in usage for the derived ones. The student LLM (\textbf{Figure~\ref{subfig:feature_frequency_Llama}}) exhibits an even stronger preference toward these original features.

This suggests that the LLMs recognize the true meaning of the tokenized features rather than treating them arbitrarily. Their implicit ability to prioritize key features over less relevant ones provides insight into LLM-driven feature selection, potentially reducing reliance on traditional methods.

\section{Broader Impact}

Our method combines LLM with a small model, and the chemical reaction produced by the teaming between the two has a significant improvement on the problem. 
This combination could also be implemented in feature selection methods~\cite{wang2024knockoff}.
In the future, we plan to expand this thinking to more areas such as business~\cite{li2023sehf,rs1,wang2024llm,rs2,rs3}, linguistics~\cite{ying2020sichuan,wang2022hierarchal}, physics law discovery~\cite{ying2025bridging}, and medicine~\cite{liu2019edta,wang2022successful,liu2024calorie,wang2024lcmdc,liu2024pth,li2024sade}.
Besides teaming, we can also apply the routing strategy~\cite{wang2025mixllm} to handle situations with multiple LLMs. The reinforcement learning strategy~\cite{ying2025bridging} and rule-based augmentation~\cite{bai2025brownian} could also be applied to enhance the robustness of the teaming method.

\end{document}